\documentclass{zgroup/zgroup}
\usepackage[utf8]{inputenc}

\usepackage{times}
\usepackage{hyperref}
\usepackage{url}

\usepackage{microtype}
\usepackage{appendix}
\usepackage{caption}
\usepackage{graphicx}
\usepackage{color}
\usepackage{booktabs}       %
\usepackage{amsfonts}       %
\usepackage{nicefrac}       %
\usepackage{multirow, multicol}
\usepackage{ragged2e}
\usepackage{amsmath}
\usepackage{thmtools, thm-restate}
\usepackage{bm, bbm}
\usepackage{booktabs} %
\usepackage{enumitem}
\usepackage{tikz}
\usepackage{wrapfig}
\usetikzlibrary{positioning}
\usepackage{lipsum}
\usepackage{float}

\usepackage{algorithm}
\usepackage{algorithmic}
\usepackage{amsmath}
\usepackage{amssymb}
\usepackage{bm}
\usepackage{bbm}
\usepackage{graphicx}
\usepackage{xcolor}

\newif\ifcomments
\commentstrue

\newcommand{\bkappa}{\bm{\kappa}}
\newcommand{\bmu}{\bm{\mu}}
\newcommand{\bomega}{\bm{\omega}}

\newcommand{\bpsi}{\bm{\psi}}
\newcommand{\btheta}{\bm{\theta}}

\newcommand{\bOmega}{\bm{\Omega}}
\newcommand{\bSigma}{\bm{\Sigma}}

\newcommand{\ff}{\mathbf{f}}

\newcommand{\xx}{\mathbf{x}}
\newcommand{\yy}{\mathbf{y}}

\renewcommand{\AA}{\mathbf{A}}

\newcommand{\KK}{\mathbf{K}}

\newcommand{\XX}{\mathbf{X}}
\newcommand{\YY}{\mathbf{Y}}

\usepackage{amsmath,amsfonts,bm}

\def\rvd{{\mathbf{d}}}

\def\rvf{{\mathbf{f}}}

\def\rvv{{\mathbf{v}}}

\def\rvx{{\mathbf{x}}}
\def\rvy{{\mathbf{y}}}
\def\rvz{{\mathbf{z}}}

\def\rvkappa{{\bm{\kappa}}}
\def\rvmu{{\bm{\mu}}}
\def\rvomega{{\bm{\omega}}}

\def\rmA{{\mathbf{A}}}

\def\rmD{{\mathbf{D}}}
\def\rmE{{\mathbf{E}}}

\def\rmK{{\mathbf{K}}}

\def\rmP{{\mathbf{P}}}

\def\rmS{{\mathbf{S}}}

\def\rmW{{\mathbf{W}}}
\def\rmX{{\mathbf{X}}}
\def\rmY{{\mathbf{Y}}}

\def\rmOmega{{\bm{\Omega}}}
\def\rmSigma{{\bm{\Sigma}}}

\DeclareMathAlphabet{\mathsfit}{\encodingdefault}{\sfdefault}{m}{sl}
\SetMathAlphabet{\mathsfit}{bold}{\encodingdefault}{\sfdefault}{bx}{n}

\title[Bayesian FSC with One-vs-Each Pólya-Gamma Augmented GPs]{Bayesian Few-Shot Classification with One-vs-Each Pólya-Gamma Augmented Gaussian Processes}
\zgroupauthor{\Name{Jake Snell}
\Email{jsnell@cs.toronto.edu}\\
\addr Vector Institute, University of Toronto \\
\Name{Richard Zemel}
\Email{zemel@cs.toronto.edu}\\
\addr Vector Institute, University of Toronto}
\date{January 2021}
\begin{document}

\maketitle

\begin{abstract}

    Few-shot classification (FSC), the task of adapting a classifier to unseen
    classes given a small labeled dataset, is an important step on the path
    toward human-like machine learning. Bayesian methods are well-suited to
    tackling the fundamental issue of overfitting in the few-shot scenario
    because they allow practitioners to specify prior beliefs and update those
    beliefs in light of observed data. Contemporary approaches to Bayesian
    few-shot classification maintain a posterior distribution over model
    parameters, which is slow and requires storage that scales with model size.
    Instead, we propose a Gaussian process classifier based on a novel
    combination of Pólya-Gamma augmentation and the one-vs-each softmax
    approximation \citep{titsias2016onevseach} that allows us to efficiently
    marginalize over functions rather than model parameters. We demonstrate
    improved accuracy and uncertainty quantification on both standard few-shot
    classification benchmarks and few-shot domain transfer tasks. 

\end{abstract}

\section{Introduction} \label{sec:introduction}

Few-shot classification (FSC) is a rapidly growing area of machine learning that
seeks to build classifiers able to adapt to novel classes given only a few
labeled examples. It is an important step towards machine learning systems that
can successfully handle challenging situations such as personalization, rare
classes, and time-varying distribution shift. The shortage of labeled data in
FSC leads to uncertainty over the parameters of the model, known as
\textit{model uncertainty} or \textit{epistemic uncertainty}. If model
uncertainty is not handled properly in the few-shot setting, there is a
significant risk of overfitting. In addition, FSC is increasingly being used for
risk-averse applications such as medical diagnosis~\citep{prabhu2019fewshot} and
human-computer interfaces \citep{wang2019customizable} where it is important for
a few-shot classifier to know when it is uncertain.

Bayesian methods maintain a distribution over model parameters and thus provide
a natural framework for capturing this inherent model uncertainty. In a Bayesian
approach, a prior distribution is first placed over the parameters of a model.
After data is observed, the posterior distribution over parameters is computed
using Bayesian inference. This elegant treatment of model uncertainty has led to
a surge of interest in Bayesian approaches to FSC that infer a posterior
distribution over the weights of a neural
network~\citep{finn2018probabilistic,yoon2018bayesian,ravi2019amortized}.

Although conceptually appealing, there are several practical obstacles to
applying Bayesian inference directly to the weights of a neural network.
Bayesian neural networks (BNNs) are expensive from both a computational and
memory perspective. Moreover, specifying meaningful priors in parameter space is
known to be difficult due to the complex relationship between weights and
network outputs~\citep{sun2019functional}.

Gaussian processes (GPs) instead maintain a distribution over \emph{functions}
rather than model parameters. The prior is directly specified by a mean and
covariance function, which may be parameterized by deep neural networks. When
used with Gaussian likelihoods, GPs admit closed form expressions for the
posterior and predictive distributions. They exchange the computational
drawbacks of BNNs for cubic scaling with the number of examples. In FSC, where
the number of examples is small, this is often an acceptable trade-off.

When applying GPs to classification with a softmax likelihood, the non-conjugacy
of the GP prior renders posterior inference intractable. Many approximate
inference methods have been proposed to circumvent this, including variational
inference and expectation propagation. In this paper we investigate a
particularly promising class of approaches that augments the GP model with a set
of auxiliary random variables, such that when the auxiliary variables are
marginalized out the original model is recovered
\citep{albert1993bayesian,girolami2006variational,linderman2015dependent}. Such
augmentation-based approaches typically admit efficient Gibbs-sampling
procedures for generating posterior samples which when combined with Fisher's
identity~\citep{douc2014nonlinear} can be used to optimize the parameters of the
mean and covariance functions.

In particular, augmentation with P\'olya-Gamma random
variables~\citep{polson2013bayesian} makes inference tractable in logistic
models. Naively, this is useful for handling binary classification, but in this
paper we show how to extend this augmentation to classification with multiple
classes by using the one-vs-each softmax approximation
\citep{titsias2016onevseach}, which can be expressed as a product of logistic
sigmoids. We further show that the one-vs-each approximation can be interpreted
as a composite likelihood~\citep{lindsay1988composite,varin2011overview}, a
connection which to our knowledge has not been made in the literature.

In this work, we make several contributions:
\begin{itemize}
  \item We show how the one-vs-each softmax approximation \citep{titsias2016onevseach}
        can be interpreted as a composite likelihood consisting of pairwise conditional terms.

  \item We propose a novel GP classification method that combines the
        one-vs-each softmax approximation with P\'olya-Gamma augmentation for tractable inference.

  \item We demonstrate competitive classification accuracy of our method on
        standard FSC benchmarks and challenging domain transfer settings.

  \item We propose several new benchmarks for uncertainty quantification in FSC,
        including calibration, robustness to input noise, and out-of-episode
        detection.

  \item We demonstrate improved uncertainty quantification of our method on the
        proposed benchmarks relative to standard few-shot baselines.
\end{itemize}

\section{Related Work}
Our work is related to both GP methods for handling non-conjugate classification
likelihoods and Bayesian approaches to few-shot classification. We summarize
relevant work here.

\subsection{GP Classification}

\paragraph{Non-augmentation approaches.} There are several classes of approaches
for applying Gaussian processes to classification. The most straightforward
method, known as least squares classification~\citep{rifkin2004defense}, treats
class labels as real-valued observations and performs inference with a Gaussian
likelihood. The Laplace approximation~\citep{williams1998bayesian} constructs a
Gaussian approximate posterior centered at the posterior mode. Variational
approaches~\citep{titsias2009variational,matthews2016sparse} maximize a lower
bound on the log marginal likelihood. In expectation
propagation~\citep{minka2001family,kim2006bayesian,hernandez-lobato2016scalable},
local Gaussian approximations to the likelihood are fitted iteratively to
minimize KL divergence from the true posterior.

\paragraph{Augmentation approaches.} Augmentation-based approaches introduce
auxiliary random variables such that the original model is recovered when
marginalized out. \citet{girolami2006variational} propose a Gaussian
augmentation for multinomial probit regression. \citet{linderman2015dependent}
utilize P\'olya-Gamma augmentation~\citep{polson2013bayesian} and a
stick-breaking construction to decompose a multinomial distribution into a
product of binomials. \citet{galy-fajou2020multiclass} propose a
logistic-softmax likelihood for classification and uses Gamma and Poisson
augmentation in addition to P\'olya-Gamma augmentation in order to perform
inference.

\subsection{Few-shot Classification}

\paragraph{Meta-learning.} A common approach to FSC is meta-learning, which seek
to learn how to update neural network parameters. The Meta-learner
LSTM~\citep{ravi2017optimization} learns a meta-level LSTM to recurrently output
a new set of parameters for the base learner. MAML~\citep{finn2017modelagnostic}
learns deep neural networks to perform well on the task-specific loss after one
or a few steps of gradient descent on the support set by directly
backpropagating through the gradient descent procedure itself.
LEO~\citep{rusu2019metalearning} performs meta-learning in a learned
low-dimensional latent space from which the parameters of a classifier are
generated.

\paragraph{Metric learning.} Metric learning approaches learn distances such
that input examples can be meaningfully compared. Siamese
Networks~\citep{koch2015siamese} learn a shared embedding network along with a
distance layer for computing the probability that two examples belong to the
same class. Matching Networks~\citep{vinyals2016matching} uses a nonparametric
classification in the form of attention over nearby examples, which can be
interpreted as a form of soft $k$-nearest neighbors in the embedding space.
Prototypical Networks~\citep{snell2017prototypical} make predictions based on
distances to nearest class centroids. Relation Networks~\citep{sung2018learning}
instead learn a more complex neural network distance function on top of the
embedding layer.

\paragraph{Bayesian Few-shot Classification.} More recently, Bayesian FSC
approaches that attempt to infer a posterior over task-specific parameters have
appeared. \citet{grant2018recasting} reinterprets MAML as an approximate
empirical Bayes algorithm and propose LLAMA, which optimizes the Laplace
approximation to the marginal likelihood. Bayesian MAML~\citep{yoon2018bayesian}
instead uses Stein Variational Gradient Descent (SVGD)~\citep{liu2016stein} to
approximate the posterior distribution over model parameters.
VERSA~\citep{gordon2019metalearning} uses amortized inference networks to obtain
an approximate posterior distribution over task-specific parameters.
ABML~\citep{ravi2019amortized} uses a few steps of Bayes by
Backprop~\citep{blundell2015weight} on the support set to produce an approximate
posterior over network parameters. CNAPs~\citep{requeima2019fast} modulate
task-specific Feature-wise Linear Modulation (FiLM)~\citep{perez2018film} layer
parameters as the output of an adaptation network that takes the support set as
input.

\paragraph{GPs for Few-shot Learning.} There have been relatively few works
applying GPs to few-shot learning. \citet{tossou2020adaptive} consider Gaussian
processes in the context of few-shot regression with Gaussian likelihoods. Deep
Kernel Transfer (DKT) \citep{patacchiola2020bayesian} use Gaussian processes
with least squares classification to perform few-shot classification and learn
covariance functions parameterized by deep neural networks. More recently,
\citet{titsias2020information} applies GPs to meta-learning with the variational
information bottleneck.

\section{Background}
In this section we review P\'olya-Gamma augmentation for binary classification
and the one-vs-each approximation before introducing our method in
Section~\ref{sec:our_method}.

\subsection{P\'olya-Gamma Augmentation}
\label{sec:binary_polya}
The P\'olya-Gamma augmentation scheme was originally introduced to address
Bayesian inference in logistic models \citep{polson2013bayesian} and has since
been applied to multinomial GPs via a stick-breaking construction
\citep{linderman2015dependent} and to GP-based classification with the logistic
softmax likelihood \citep{galy-fajou2020multiclass}.

Suppose we have a vector of logits $\bpsi \in \mathbb{R}^N$ with corresponding
binary labels $\yy \in \{0,1\}^N$. The logistic likelihood is
\begin{equation}
  p(\yy | \bpsi) = \prod_{i=1}^N \sigma(\psi_i)^{y_i} (1 - \sigma(\psi_i))^{1 - y_i} = \prod_{i=1}^N \frac{(e^{\psi_i})^{y_i}}{1 + e^{\psi_i}},
  \label{eq:binary_likelihood}
\end{equation}
where $\sigma(\cdot)$ is the logistic sigmoid function. Let the prior over
$\bpsi$ be Gaussian: $p(\bpsi) = \mathcal{N}(\bpsi | \bmu, \bSigma).$ In
Bayesian inference, we are interested in the posterior
$p(\bpsi | \yy) \propto p(\yy | \bpsi) p(\bpsi)$ but the form of
\eqref{eq:binary_likelihood} does not admit analytic computation of the
posterior due to non-conjugacy. The main idea of P\'olya-Gamma augmentation is
to introduce auxiliary random variables $\bomega$ to the likelihood such that
the original model is recovered when $\bomega$ is marginalized out:
$p(\yy | \bpsi) = \int p(\bomega) p(\yy | \bpsi, \bomega) \, d\bomega$. The
P\'olya-Gamma distribution $\omega \sim \text{PG}(b, c)$ can be written as an
infinite convolution of Gamma distributions:
\begin{equation}
  \omega \overset{D}{=} \frac{1}{2\pi^2} \sum_{k=1}^\infty \frac{\text{Ga}(b, 1)}{ (k - 1/2)^2 + c^2 / (4 \pi^2) }.
\end{equation}
The following integral identity holds for $b>0$:
\begin{equation}
  \frac{ (e^\psi)^a }{ (1 + e^\psi)^b} = 2^{-b} e^{\kappa \psi} \int_0^\infty e^{-\omega \psi^2  /2} p(\omega) \, d\omega,
  \label{eq:polya_integral_identity}
\end{equation}
where $\kappa = a - b/2$ and $\omega \sim \text{PG}(b, 0).$ Specifically, when
$a = y$ and $b = 1$, we recover an individual term of the logistic likelihood
\eqref{eq:binary_likelihood}:
\begin{equation}
  p(y | \psi) = \frac{ (e^{\psi})^{y} }{1 + e^{\psi}} = \frac{1}{2} e^{\kappa \psi} \int_0^\infty e^{-\omega \psi^2/2} p(\omega) \, d\omega,
\end{equation}
where $\kappa = y - 1/2$ and $\omega \sim PG(1, 0)$. Conditioned on $\bomega,$
the batch likelihood is proportional to a diagonal Gaussian:
\begin{equation} \label{eq:augmented_binary_likelihood} p(\yy | \bpsi, \bomega) \propto \prod_{i=1}^N e^{-\omega_i \psi_i^2/2} e^{\kappa_i \psi_i} \propto \mathcal{N}(\bOmega^{-1} \bkappa \,|\, \bpsi, \bOmega^{-1}),
\end{equation}
where $\kappa_i = y_i - 1/2$ and $\bOmega = \text{diag}(\bomega)$. The
conditional distribution over $\bpsi$ given $\yy$ and $\bomega$ is now
tractable:
\begin{equation}
  \label{eq:binary_psi_posterior}
  p(\bpsi | \yy, \bomega) \propto p(\yy | \bpsi, \bomega) p(\bpsi) \propto \mathcal{N}(\bpsi | \tilde{\bSigma} (\bSigma^{-1} \bmu + \bkappa), \tilde{\bSigma}),
\end{equation}
where $\tilde{\bSigma} = (\bSigma^{-1} + \bOmega)^{-1}$. The conditional
distribution of $\bomega$ given $\bpsi$ and $\yy$ can also be easily computed:
\begin{equation} \label{eq:binary_pg_sample} p(\omega_i | y_i, \psi_i) \propto \text{PG}(\omega_i | 1, 0) e^{-\omega_i \psi^2_i / 2} \propto \text{PG}(\omega_i | 1, \psi_i),
\end{equation}
where the last expression follows from the exponential tilting property of
P\'olya-Gamma random variables. This suggest a Gibbs sampling procedure in which
iterates $\bomega^{(t)} \sim p(\bomega | \yy, \bpsi^{(t-1)})$ and
$\bpsi^{(t)} \sim p(\bpsi | \XX, \yy, \bomega^{(t)})$ are drawn sequentially
until the Markov chain reaches its stationary distribution, which is the joint
posterior $p(\bpsi, \bomega | \yy)$. Fortunately, efficient samplers for the
P\'olya-Gamma distribution have been developed \citep{windle2014sampling} to
facilitate this.

\subsection{One-vs-Each Approximation to Softmax}

The one-vs-each (OVE) approximation \citep{titsias2016onevseach} was formulated
as a lower bound to the softmax likelihood in order to handle classification
over a large number of output classes, where computation of the normalizing
constant is prohibitive. We employ the OVE approximation not to deal with
extreme classification, but rather due to its compatibility with P\'olya-Gamma
augmentation, as we shall soon see. The one-vs-each approximation can be derived
by first rewriting the softmax likelihood as follows:
\begin{equation}
  p(y = i \,|\, \ff) \triangleq \frac{e^{f_i}}{\sum_{j} e^{f_{j}}} = \frac{1}{1 + \sum_{j\neq i} e^{-(f_i - f_{j})}},
  \label{eq:softmax_rewrite}
\end{equation}
where $\ff \triangleq (f_1, \ldots, f_C)^\top$ are the logits. Since in general
$\prod_k (1 + \alpha_k) \ge (1 + \sum_k \alpha_k)$ for $\alpha_k \ge 0$, the
softmax likelihood \eqref{eq:softmax_rewrite} can be bounded as follows:
\begin{equation}
  p(y = i \,|\, \mathbf{f}) \ge \prod_{j \neq i} \frac{1}{1 + e^{-(f_i - f_{j})}} = \prod_{j \neq i} \sigma(f_i - f_{j}),
  \label{eq:ove_bound}
\end{equation}
which is the OVE lower bound. This expression avoids the normalizing constant
and factorizes into a product of pairwise sigmoids, which is amenable to
P\'olya-Gamma augmentation for tractable inference.

\section{One-vs-Each P\'olya-Gamma GPs}\label{sec:our_method} In this section we
introduce our method for GP-based Bayesian few-shot classification, which brings
together P\'olya-Gamma augmentation and the one-vs-each (OVE) approximation in a
novel combination. In Section~\ref{sec:ove_composite}, we show how the OVE
approximation can be interpreted as a pairwise composite likelihood. In
Section~\ref{sec:ove_gp_classification}, we discuss the use of OVE as a
likelihood for GP classification. In Section~\ref{sec:gp_inference_gibbs}, we
show how OVE combined with P\'olya-Gamma augmentation enables tractable
inference with Gibbs sampling. Then in Section~\ref{sec:posterior_predictive} we
show how to compute the posterior predictive distribution using the Gibbs
output. In Section~\ref{sec:polya_few_shot}, we describe how we learn the
covariance hyperparameters with marginal likelihood and predictive likelihood
objectives. Finally, in Section~\ref{sec:kernel_choice_main}, we discuss our
choice of covariance function.

\subsection{OVE as a Composite Likelihood}\label{sec:ove_composite}

\citet{titsias2016onevseach} showed that the OVE approximation shares the same
global optimum as the softmax maximum likelihood, suggesting a close
relationship between the two. We show here that in fact OVE can be interpreted
as a pairwise \emph{composite likelihood} version of the softmax. Composite
likelihoods~\citep{lindsay1988composite,varin2011overview} are a type of
approximate likelihood often employed when the exact likelihood is intractable
or otherwise difficult to compute. Given a collection of marginal or conditional
events $\{E_{1}, \ldots, E_{K}\}$ and parameters $\ff$, a composite likelihood
is defined as:
\begin{equation}
  \mathcal{L}_{\text{CL}}(\ff \,|\, y) \triangleq \prod_{k=1}^K \mathcal{L}_k(\ff \,|\, y)^{w_k},
\end{equation}
where $\mathcal{L}_{k}(\ff \,|\, y) \propto p(y \in E_{k} \,|\, \ff)$ and
$w_{k} \ge 0$ are arbitrary weights.

In order to make the connection to OVE, it will be useful to let the one-hot
encoding of the label $y$ be denoted as $\yy \in \{0, 1\}^{C}$. Define a set of
$C(C-1)/2$ pairwise conditional events $E_{ij}$, one each for all pairs of
classes $i \neq j$, indicating the event that the model's output matches the
target label for classes $i$ and $j$ conditioned on all the other classes:
\begin{equation}\label{eq:pairevents}
  p(\yy \in E_{{ij}} \,|\, \ff) \triangleq p(y_{i}, y_{j} \,|\, \yy_{\neg ij}, \ff),
\end{equation}
where $\neg ij$ denotes the set of classes not equal to either $i$ or $j$. This
expression resembles the pseudolikelihood~\citep{besag1975statistical}, but
instead of a single conditional event per output site, the expression in
\eqref{eq:pairevents} considers all pairs of sites.
\citet{stoehr2015calibration} explored similar composite likelihood
generalizations of the pseudolikelihood in the context of random fields.

Now suppose that $y_{c} = 1$ for some class $c \notin \{i,j\}$. Then
$p(y_{i}, y_{j} \,|\, \yy_{\neg ij}, \ff) = 1$ due to the one-hot constraint.
Otherwise either $y_{i} = 1$ or $y_{j} = 1$. In this case, assume without loss
of generality that $y_{i} = 1$ and $y_{j} = 0$ and thus
\begin{equation}
  p(y_{i}, y_{j} \,|\, \yy_{\neg ij}, \ff) = \frac{e^{f_{i}}}{e^{f_{i}} + e^{f_{j}}} = \sigma(f_{i} - f_{j}).
\end{equation}
The composite likelihood defined in this way with unit component weights is
therefore
\begin{equation}\label{eq:ovecomposite}
  \mathcal{L}_{\text{OVE}}(\ff \,|\, \yy) = \prod_{i} \prod_{j \neq i} p(y_{i}, y_{j} | \yy_{\neg ij}, \ff) = \prod_{i} \prod_{j \neq i} \sigma(f_{i} - f_{j})^{y_{i}}.
\end{equation}
Alternatively, we may simply write
$\mathcal{L}^{\text{OVE}}(\ff \,|\, y = i) = \prod_{j \neq i} \sigma(f_{i} - f_{j})$,
which is identical to the OVE bound~\eqref{eq:ove_bound}.

\subsection{GP Classification with the OVE
  Likelihood}\label{sec:ove_gp_classification}

We now turn our attention to GP classification. Suppose we have access to
examples $\XX \in \mathbb{R}^{N \times D}$ with corresponding one-hot labels
$\YY \in \{0,1\}^{N \times C}$, where $C$ is the number of classes. We consider
the logits jointly as a single vector
\begin{equation}
  \ff \triangleq (f_1^1, \ldots, f_N^1, f_1^2, \ldots, f_N^2, \ldots, f_1^C, \ldots, f_N^C)^\top
\end{equation}
and place an independent GP prior on the logits for each class:
$\ff^c(\xx) \sim \mathcal{GP}(m(\xx), k(\xx, \xx'))$. Therefore we have
$p(\ff | \XX) = \mathcal{N}(\ff | \bmu, \KK),$ where $\mu_i^c = m(\xx_i)$ and
$\KK$ is block diagonal with $K_{ij}^c = k(\xx_i, \xx_j)$ for each block
$\KK^c$.

The P\'olya-Gamma integral identity used to derive
\eqref{eq:augmented_binary_likelihood} does not have a multi-class analogue and
thus a direct application of the augmentation scheme to the softmax likelihood
is nontrivial. Instead, we propose to directly replace the softmax with the
OVE-based composite likelihood function from \eqref{eq:ovecomposite} with unit
weights:
\begin{equation}
  \mathcal{L}^{\text{OVE}}(\ff_{i} | y_{i} = c) \triangleq \prod_{{c' \neq c}} \sigma(f_{i}^{c} - f_{i}^{c'}).
  \label{eq:ove_likelihood}
\end{equation}

The posterior over $\ff$ when using \eqref{eq:ove_likelihood} as the likelihood
function can therefore be expressed as:
\begin{equation}
  p(\ff | \XX, \yy) \propto p(\ff | \XX) \prod_{i=1}^{N}  \prod_{c' \neq y_{i}} \sigma(f_{i}^{y_{i}} - f_{i}^{c'}),
\end{equation}
to which P\'olya-Gamma augmentation can be applied as we show in the next
section. Our motivation for using a composite likelihood therefore differs from
the traditional motivation, which is to avoid the use of a likelihood function
which is intractable to \emph{evaluate}. Instead, we employ a composite
likelihood because it makes posterior \emph{inference} tractable when coupled
with P\'olya-Gamma augmentation.

Prior work on Bayesian inference with composite likelihoods has shown that the
composite posterior is consistent under fairly general conditions
\citep{miller2019asymptotic} but can produce overly concentrated
posteriors~\citep{pauli2011bayesian,ribatet2012bayesian} because each component
likelihood event is treated as independent when in reality there may be
significant dependencies. Nevertheless, we show in Section~\ref{sec:experiments}
that in practice our method exhibits competitive accuracy and strong calibration
relative to baseline few-shot learning algorithms. We leave further theoretical
analysis of the OVE composite posterior and its properties for future work.

Compared to choices of likelihoods used by previous approaches, there are
several reasons to prefer OVE. Relative to the Gaussian augmentation approach of
\citet{girolami2006variational}, P\'olya-Gamma augmentation has the benefit of
fast mixing and the ability of a single value of $\bomega$ to capture much of
the marginal distribution over function values\footnote{See in particular
  Appendix C of \citep{linderman2015dependent} for a detailed explanation of
  this phenomenon.}. The stick-breaking construction of
\citet{linderman2015dependent} induces a dependence on the ordering of classes,
which leads to undesirable asymmetry. Finally, the logistic-softmax likelihood
of \citet{galy-fajou2020multiclass} requires three augmentations and careful
learning of the mean function to avoid \textit{a priori} underconfidence (see
Section~\ref{sec:lsm_calibration} for more details).

\subsection{Posterior Inference via Gibbs Sampling}\label{sec:gp_inference_gibbs}
We now describe how we perform tractable posterior inference in our model with
Gibbs sampling. Define the matrix
$\AA \triangleq \small{\textsc{OVE-MATRIX}}(\YY)$ to be a $CN \times CN$ sparse
block matrix with $C$ row partitions and $C$ column partitions. Each block
$\AA_{cc'}$ is a diagonal $N \times N$ matrix defined as follows:
\begin{equation}
  \AA_{cc'} \triangleq \text{diag}(\YY_{\cdot c'}) - \mathbbm{1}[c = c']
  \mathbf{I}_n,
\end{equation}
where $\YY_{\cdot c'}$ denotes the $c'$th column of $\mathbf{Y}$. Now the binary
logit vector $\bpsi \triangleq \AA \ff \in \mathbb{R}^{CN}$ will have entries
equal to $f_i^{y_i} - f_i^{c}$ for each unique combination of $c$ and $i,$ of
which there are $CN$ in total. The OVE composite likelihood can now be written
as $\mathcal{L}(\bpsi | \YY) = 2^N \prod_{j=1}^{NC} \sigma(\psi_j)$, where the
$2^N$ term arises from the $N$ cases in which $\psi_j = 0$ due to comparing the
ground truth logit with itself.

Analogous to \eqref{eq:augmented_binary_likelihood}, the likelihood of $\bpsi$
conditioned on $\bomega$ and $\YY$ is proportional to a diagonal Gaussian:
\begin{equation}
  \mathcal{L}(\bpsi | \YY, \bomega) \propto \prod_{j=1}^{NC} e^{-\omega_j \psi_j^2/2}e^{\kappa_j \psi_j} \propto \mathcal{N}(\bOmega^{-1} \bkappa |\bpsi , \bOmega^{-1}),
  \label{eq:psi_likelihood}
\end{equation}
where $\kappa_j = 1/2$ and $\bOmega = \text{diag}(\bomega)$. By exploiting the
fact that $\bpsi = \AA \ff,$ we can express the likelihood in terms of $\ff$ and
write down the conditional composite posterior as follows:
\begin{equation}
  \label{eq:f_conditional}
  p(\ff | \XX, \YY, \bomega) \propto \mathcal{N}(\bOmega^{-1} \bkappa | \AA \ff, \bOmega^{-1}) \mathcal{N}(\ff | \bmu, \KK) \propto \mathcal{N}(\ff | \tilde{\bSigma} (\KK^{-1} \bmu + \AA^\top \bkappa), \tilde{\bSigma}),
\end{equation}
where $\tilde{\bSigma} = (\KK^{-1} + \AA^\top \bOmega \AA)^{-1}$, which is an
expression remarkably similar to \eqref{eq:binary_psi_posterior}. Analogous to
\eqref{eq:binary_pg_sample}, the conditional distribution over $\bomega$ given
$\ff$ and the data becomes
$p(\bomega | \yy, \ff) = \text{PG}(\bomega | \bm{1}, \AA \ff)$.

The primary computational bottleneck of posterior inference lies in sampling
$\ff$ from \eqref{eq:f_conditional}. Since $\tilde{\bSigma}$ is a $CN \times CN$
matrix, a naive implementation has complexity $\mathcal{O}(C^3 N^3)$. By
utilizing the matrix inversion lemma and Gaussian sampling techniques
summarized in \citep{doucet2010note}, this can be brought down to
$\mathcal{O}(C N^3).$ Details may be found in
Section~\ref{sec:efficient_sampling}.

\subsection{Posterior Predictive Distribution}\label{sec:posterior_predictive}
The posterior predictive distribution for a query example $\xx_*$ conditioned on
$\bomega$ is:
\begin{equation}
  \label{eq:posterior_predictive} p(\yy_* | \xx_*, \XX, \YY, \bomega) = \int p(\yy_* | \ff_*) p(\ff_* | \xx_*, \XX, \YY, \bomega)\, d\ff_*,
\end{equation}
where $\ff_*$ are the query example's logits. The predictive distribution over
$\ff_*$ can be obtained by noting that $\bpsi$ and the query logits are jointly
Gaussian:
\begin{equation}
  \left[
    \begin{array}{c} \bpsi \\ \ff_* \end{array}
  \right] \sim \mathcal{N}\left( 0,
    \left[
      \begin{array}{cc} \AA \KK \AA^\top + \bOmega^{-1} & \AA \KK_* \\  (\AA \KK_*)^\top & \KK_{**} \end{array}
    \right] \right),
\end{equation}
where $\KK_*$ is the $NC \times C$ block diagonal matrix with blocks
$K_\theta(\XX, \xx_*)$ and $\KK_{**}$ is the $C \times C$ diagonal matrix with
diagonal entries $k_\theta(\xx_*, \xx_*)$. The predictive distribution becomes:
\begin{align} \begin{split}
    &p(\ff_* | \xx_*, \XX, \YY, \bomega) = \mathcal{N}(\ff_* | \bmu_*, \bSigma_*), \text{ where} \\
    &\bmu_* =  (\AA \KK_*)^\top (\AA \KK \AA^\top + \bOmega^{-1})^{-1} \bOmega^{-1} \bkappa \text{ and} \\
    &\bSigma_* = \KK_{**} - (\AA \KK_*)^\top (\AA \KK \AA^\top + \bOmega^{-1})^{-1} \AA \KK_*.
  \end{split} \end{align} With $p(\ff_* | \xx_*, \XX, \YY, \bomega)$ in hand,
the integral in \eqref{eq:posterior_predictive} can easily be computed
numerically for each class $c$ by forming the corresponding OVE linear
transformation matrix $\AA^c$ and then performing 1D Gaussian-Hermite quadrature
on each dimension of
$\mathcal{N}(\bpsi_*^c | \AA^c \bmu^*, \AA^c \bSigma_* \AA^{c\top})$.

\subsection{Learning Covariance Hyperparameters for Few-shot Classification}\label{sec:polya_few_shot}
We now describe how we apply OVE P\'olya-Gamma augmented GPs to few-shot
classification. We assume the standard episodic few-shot setup in which one
observes a labeled support set $\mathcal{S} = (\XX, \YY)$. Predictions must then
be made for a query example $(\xx_*, \yy_*)$.

We consider a zero-mean GP prior over the class logits
$\ff^c(\xx) \sim \mathcal{GP}(\bm{0}, k_{\btheta}(\xx, \xx')),$ where $\btheta$
are learnable parameters of our covariance function. These could include
traditional hyperparameters such as lengthscales or the weights of a deep neural
network as in deep kernel learning \citep{wilson2016deep}. By performing
Bayesian modeling on the logits directly, we are able to construct a posterior
distribution over functions and use it to make predictions on the query
examples. The reader is encouraged to refer to
\citep[Section~2.2]{rasmussen2006gaussian} for a discussion on the
correspondence between function-space and weight-space.

We consider two objectives for learning hyperparameters of the covariance
function: the marginal likelihood and the predictive likelihood. Marginal
likelihood measures the likelihood of the hyperparameters given the observed
data and is intuitively appealing from a Bayesian perspective. On the other
hand, many standard FSC methods optimize for predictive likelihood on the query
set \citep{vinyals2016matching,finn2017modelagnostic,snell2017prototypical}.
Both objectives marginalize over latent functions, thereby making full use of
our Bayesian formulation.

\noindent \textbf{Marginal Likelihood (ML).} The log marginal likelihood can be
written as follows:
\begin{align}\label{eq:marginal_likelihood}
  L_{\text{ML}}(\btheta; \XX, \YY) \triangleq \log p_{\bm{\theta}}(\YY | \XX) &= \log \int p(\bomega) p_{\bm{\theta}}(\YY| \bomega, \XX) \, d\bomega \nonumber \\
                                                                              &= \log \int p(\bomega) \int  \mathcal{L}(\ff | \YY, \bomega) p_{\bm{\theta}}(\ff | \XX)\,d\ff \, d\bomega
\end{align}
The gradient of the log marginal likelihood can be estimated by posterior
samples $\bomega \sim p_{\btheta}(\bomega | \XX, \YY).$ In practice, we use a
stochastic training objective based on samples of $\bomega$ from Gibbs chains.
We use Fisher's identity \citep{douc2014nonlinear} to derive the following
gradient estimator:
\begin{equation}
  \nabla_{\btheta} L_{\text{ML}} = \int p_{\btheta}(\bomega | \XX, \YY) \nabla_{\btheta} \log p_{\bm{\theta}}(\YY | \bomega, \XX) \, d\bomega
  \approx \frac{1}{M} \sum_{m=1}^M \nabla_{\btheta} \log p_{\btheta}(\YY | \XX, \bomega^{(m)}),
\end{equation}
where $\bomega^{(1)}, \ldots, \bomega^{(M)}$ are samples from the posterior
Gibbs chain. As suggested by \citet{patacchiola2020bayesian}, who applied GPs to
FSC via least-squares classification, we merge the support and query sets during
learning to take full advantage of the available data within each episode.

\noindent \textbf{Predictive Likelihood (PL).} The log predictive likelihood for
a query example $\xx_*$ is:
\begin{align}
  L_\text{PL}(\btheta; \XX, \YY, \xx_{*}, \yy_{*}) \triangleq \log p_{\btheta}(\yy_* | \xx_*, \XX, \YY) =  \log \int p(\bomega) p_{\btheta}(\yy_* | \xx_*, \XX, \YY, \bomega) \, d\bomega.
\end{align}
We use an approximate gradient estimator again based on posterior samples of
$\bomega$:
\begin{equation}
  \nabla_{\btheta} L_\text{PL} \approx  \int p_{\btheta}(\bomega | \XX, \YY) \nabla_{\btheta} \log p_{\btheta}(\yy_* | \xx_*, \XX, \YY)\, d\bomega  \approx \frac{1}{M} \sum_{m=1}^M \nabla_{\btheta} \log p_{\btheta}(\yy_* | \xx_*, \XX, \YY, \bomega^{(m)}).
\end{equation}
We note that this is not an unbiased estimator of the gradient, but find it
works well in practice. Our learning algorithm for both marginal and predictive
likelihood is summarized in Algorithm~\ref{alg:pg_learning}.

\begin{algorithm}[!htb]
  \caption{One-vs-Each P\'olya-Gamma GP Learning}
  \label{alg:pg_learning}
  \begin{algorithmic}
    \STATE {\bfseries Input:} Objective
    $L \in \{ L_{\text{ML}}, L_{\text{PL}} \}$, Task distribution $\mathcal{T}$,
    number of parallel Gibbs chains $M$, number of steps $T$, learning rate
    $\eta$. \vspace{.1cm} \STATE Initialize hyperparameters $\btheta$ randomly.
    \REPEAT \STATE Sample
    $\mathcal{S} = (\XX, \YY), \mathcal{Q} = (\XX_*, \YY_*) \sim \mathcal{T}$
    \IF {$L = L_{\text{ML}}$} \STATE
    $\XX \leftarrow \XX \cup \XX_*$, $\YY \leftarrow \YY \cup \YY_*$ \ENDIF
    \STATE $\AA \leftarrow \small{\textsc{OVE-MATRIX}}(\YY)$ \FOR{$m=1$
      {\bfseries to} $M$} \STATE $\bomega_0^{(m)} \sim PG(1, 0)$,
    $\ff_0^{(m)} \sim p_{\btheta}(\ff | \XX)$ \FOR{$t=1$ {\bfseries to} $T$}
    \STATE $\bpsi_t^{(m)} \leftarrow \AA \ff^{(m)}_{t-1} $ \STATE
    $\bomega_t^{(m)} \sim \text{PG}(1, \bpsi_t^{(m)})$ \STATE
    $\ff_t^{(m)} \sim p_{\btheta}(\ff | \XX, \YY, \bomega_t^{(m)})$ \ENDFOR
    \ENDFOR \IF {$L = L_{\text{ML}}$} \STATE
    $\btheta \leftarrow \btheta + \frac{\eta}{M} \sum_{m=1}^M \nabla_{\btheta} \log p_{\btheta}(\YY | \XX, \bomega^{(m)}_T)$
    \ELSE \STATE
    $\btheta \leftarrow \btheta + \frac{\eta}{M} \sum_{m=1}^M \sum_j \nabla_{\btheta} \log p_{\btheta}(\yy_{*j} | \xx_{*j}, \mathcal{S}, \bomega^{(m)}_T)$
    \ENDIF

    \UNTIL{convergence}
  \end{algorithmic}
\end{algorithm}

\subsection{Choice of Covariance Function}\label{sec:kernel_choice_main}
For our method we primarily use the following covariance function, which we
refer to as the ``cosine'' kernel due to its similarity to cosine similarity:
\begin{equation}\label{eq:cosine_kernel}
  k^\text{cos}(\xx, \xx'; \btheta, \alpha) = \exp(\alpha) \frac{g_{\btheta}(\xx)^\top g_{\btheta}(\xx')}{\|g_{\btheta}(\xx)\| \|g_{\btheta}(\xx')\|},
\end{equation}
where $g_{\btheta}(\cdot)$ is a deep neural network that outputs a
fixed-dimensional encoded representation of the input and $\alpha$ is the scalar
log output scale. We experimented with several kernels and found the cosine and
linear kernels to generally outperform RBF-based kernels (see
Section~\ref{sec:kernel_choice} for detailed comparisons). We hypothesize that
this is because they help the embedding network $g_{\btheta}(\cdot)$ to learn
linearly separable representations. In contrast, the RBF-based kernels yields
nonlinear decision boundaries with respect to the embedded representation and
may not provide the embedding network with as strong of a learning signal.
Further study of the benefits and drawbacks of linear vs. nonlinear kernels is
an interesting area of future work.

\section{Experiments}\label{sec:experiments}

In Section~\ref{sec:likelihood_comparison}, we compare the OVE likelihood to
alternative likelihoods on toy examples in order to gain intuition about its
strengths and weaknesses. We then proceed to evaluate the performance of our
method on few-shot classification in
Sections~\ref{sec:fewshot_experiment_overview}-\ref{sec:fewshot_ooe}.

\subsection{Comparison of Likelihoods}\label{sec:likelihood_comparison}

In this section we seek to better understand the behaviors of the softmax, OVE,
logistic softmax, and Gaussian likelihoods for classification. For convenience,
we summarize the forms of these likelihoods in Table~\ref{tab:likelihoodrecap}.

\begin{table}[!ht]
  \centering \renewcommand{\arraystretch}{2}
  \caption{Likelihoods used in
    Section~\ref{sec:likelihood_comparison}.}\label{tab:likelihoodrecap}
  \begin{tabular}{lll}
    \toprule
    \bf{Likelihood} & $\mathcal{L}(\rvf \,|\, y = c)$ \\ \midrule
    Softmax & $\displaystyle \frac{\exp(f_c)}{\sum_{c'} \exp(f_{c'})}$ \\
    Gaussian & $\displaystyle \prod_{c'} \mathcal{N}( 2 \cdot \mathbbm{1}[c' = c] - 1 \,|\, \mu = f_{c'}, \sigma^2 = 1)$ \\
    Logistic Softmax (LSM) & $\displaystyle \frac{\sigma(f_c)}{\sum_{c'} \sigma(f_{c'})}$ \\
    One-vs-Each (OVE) & $\displaystyle \prod_{c' \neq c} \sigma(f_c - f_{c'})$ \\ \bottomrule
  \end{tabular}
\end{table}

\subsubsection{Histogram of Confidences}\label{sec:lsm_calibration}

\begin{figure*}[htb]
  \begin{center}
    \centerline{\includegraphics[width=0.7\textwidth]{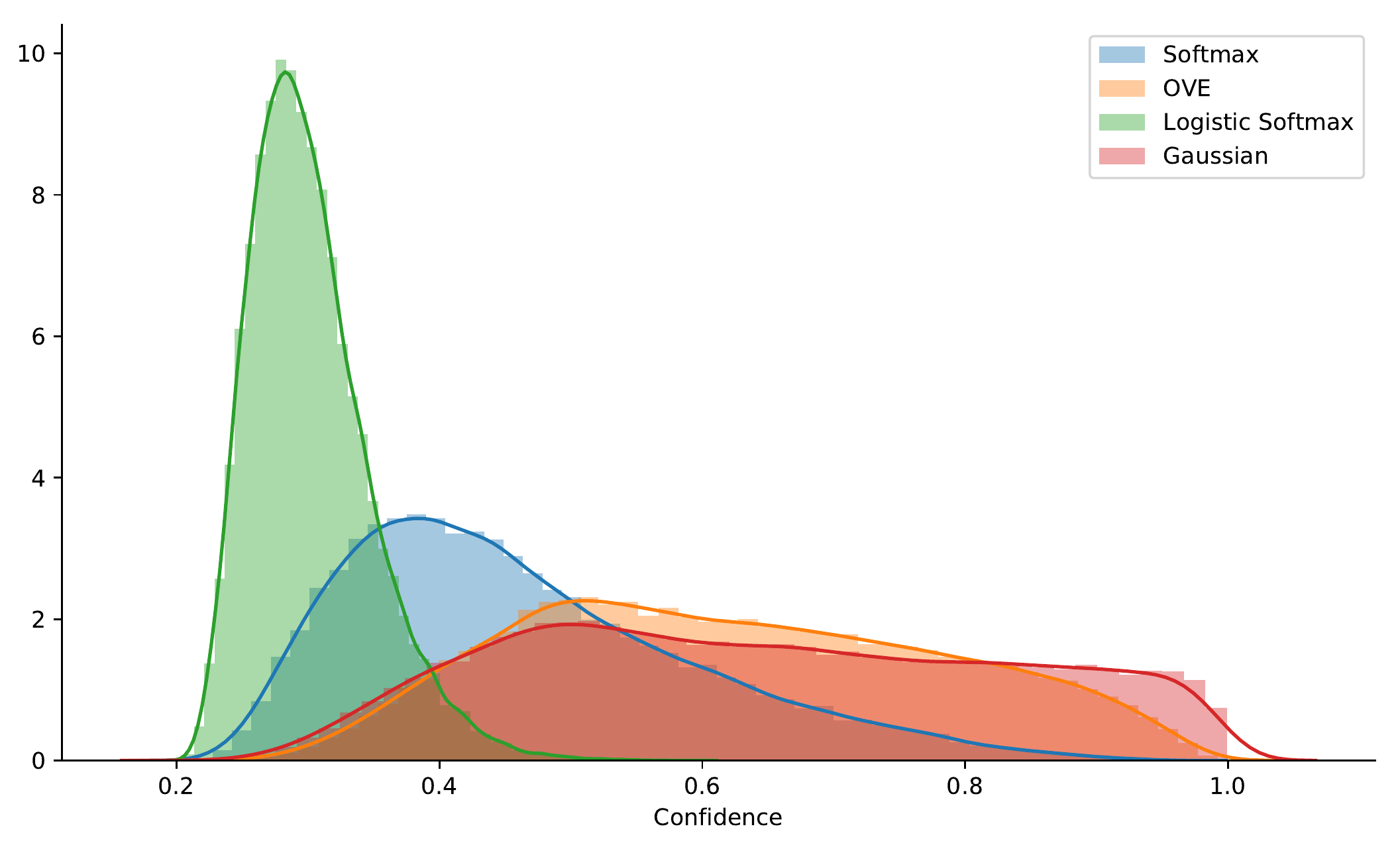}}
    \caption{Histogram and kernel density estimate of confidence for randomly
      generated function samples $f_{c} \sim \mathcal{N}(0, 1)$. Normalized
      output probabilities were computed for $C=5$ and a histogram of
      $\max_c p(y=c|\mathbf{f})$ was computed for 50,000 randomly generated
      simulations.}\label{fig:logit_simulation}
  \end{center}
\end{figure*}

We sampled logits from $f_{c} \sim \mathcal{N}(0, 1)$ and plotted a histogram
and kernel density estimate of the maximum output probability
$\max_c p(y=c \,|\, \mathbf{f})$ for each of the likelihoods shown in
Table~\ref{tab:likelihoodrecap}, where $C = 5$. The results are shown in
Figure~\ref{fig:logit_simulation}. Logistic softmax is \textit{a priori}
underconfident: it puts little probability mass on confidence above 0.4. This
may be due to the use of the sigmoid function which squashes large values of
$f$. Gaussian likelihood and OVE are \textit{a priori} overconfident in that
they put a large amount of probability mass on confident outputs. Note that this
is not a complete explanation, because GP hyperparameters such as the prior mean
or Gaussian likelihood variance may be able to compensate for these
imperfections to some degree. Indeed, we found it helpful to learn a constant
mean for the logistic softmax likelihood, as mentioned in
Section~\ref{sec:baselines}.

\subsubsection{Likelihood Visualization}\label{sec:likelihood_vis}

\begin{figure}[!ht]
  \centering
  \subfigure[Softmax]{\includegraphics[width=0.24\linewidth]{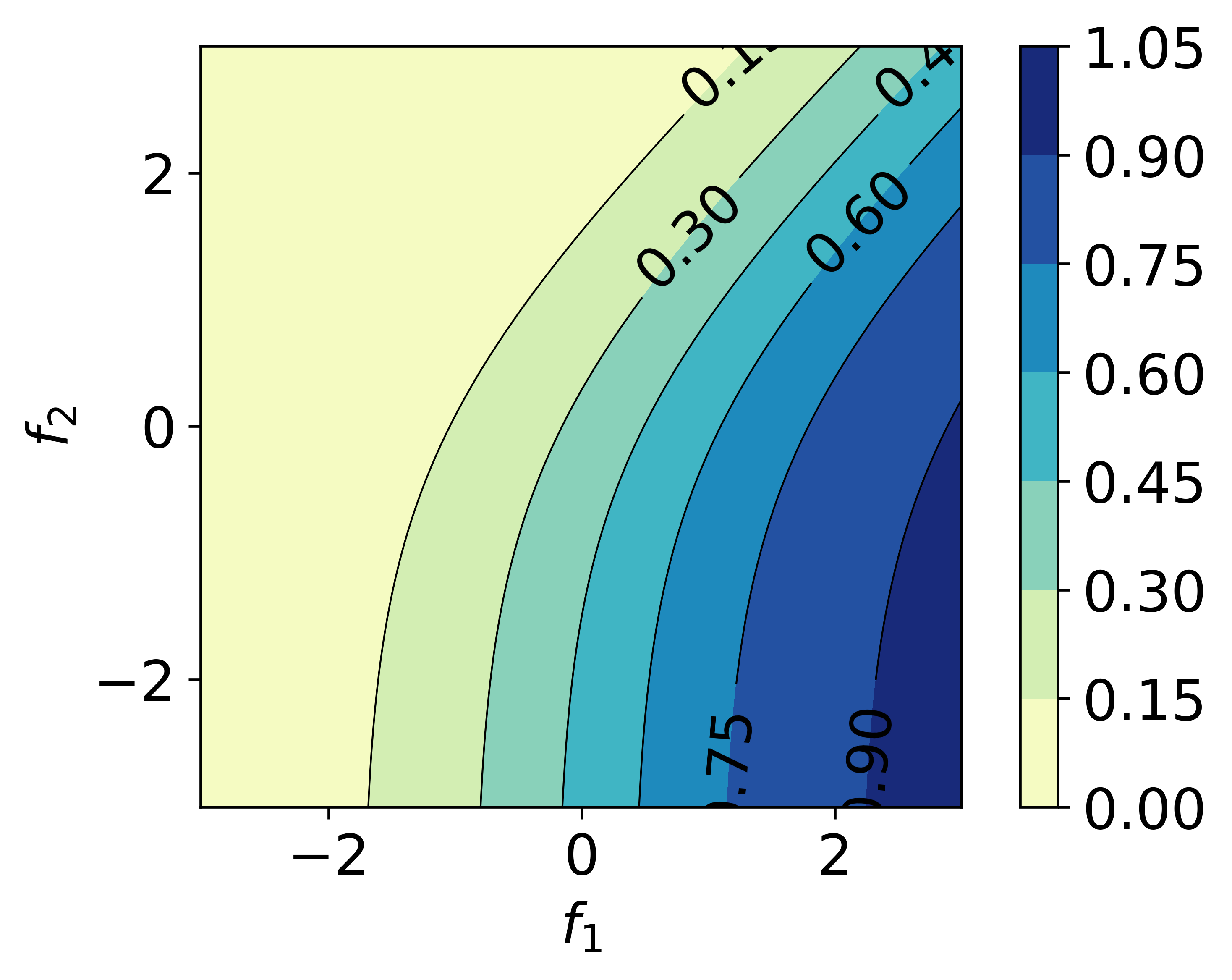}}
  \subfigure[Gaussian]{\includegraphics[width=0.24\linewidth]{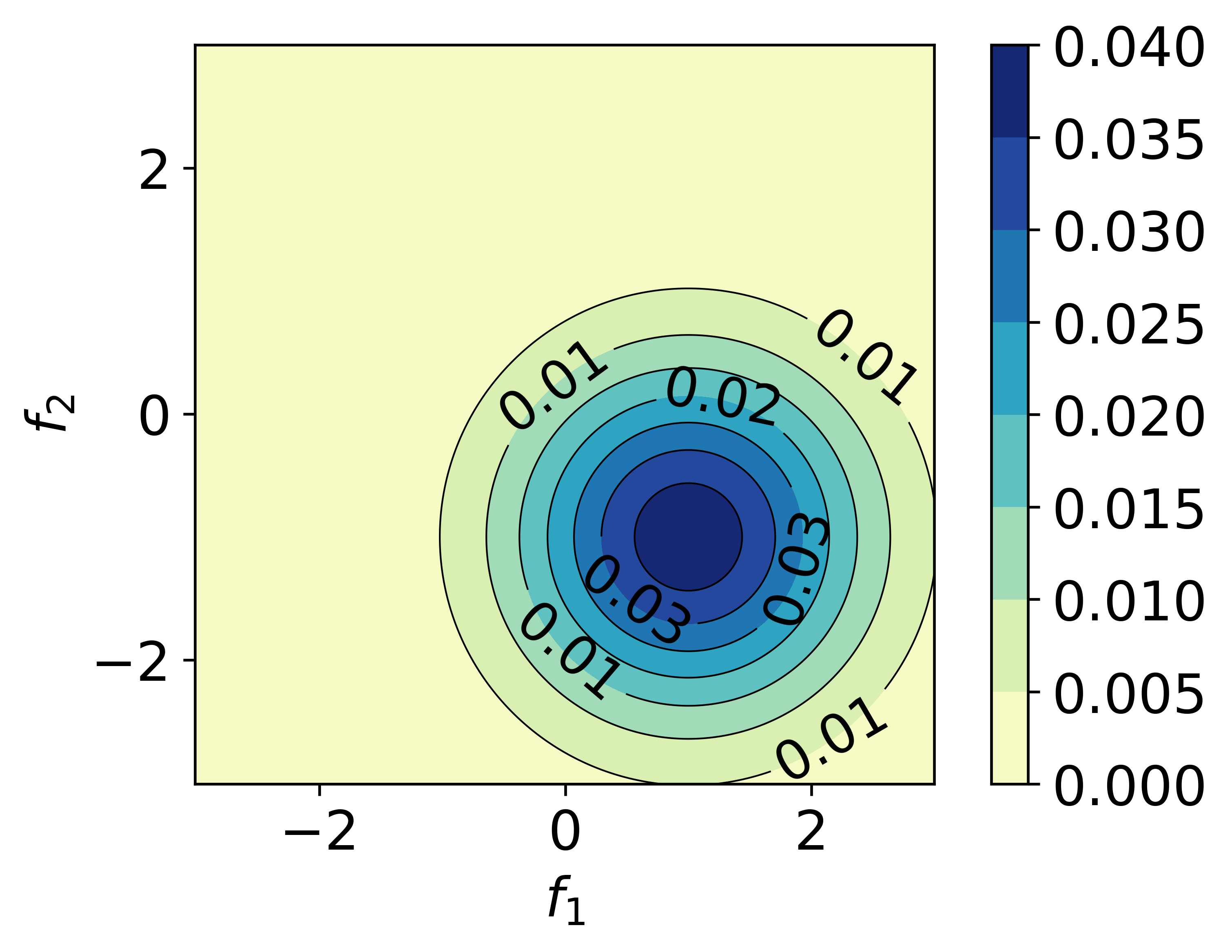}}
  \subfigure[Logistic
  Softmax]{\includegraphics[width=0.24\linewidth]{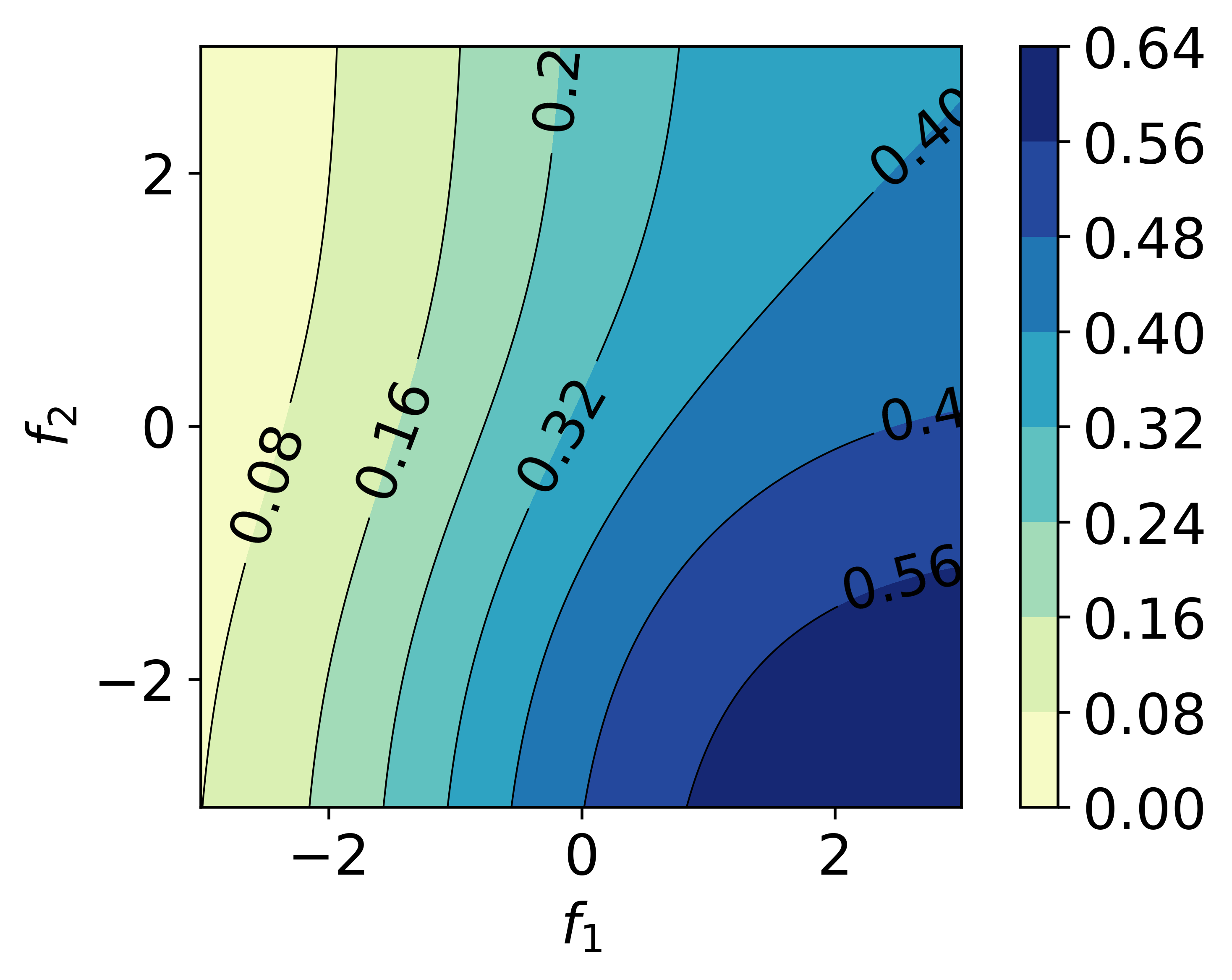}}
  \subfigure[One-vs-Each]{\includegraphics[width=0.24\linewidth]{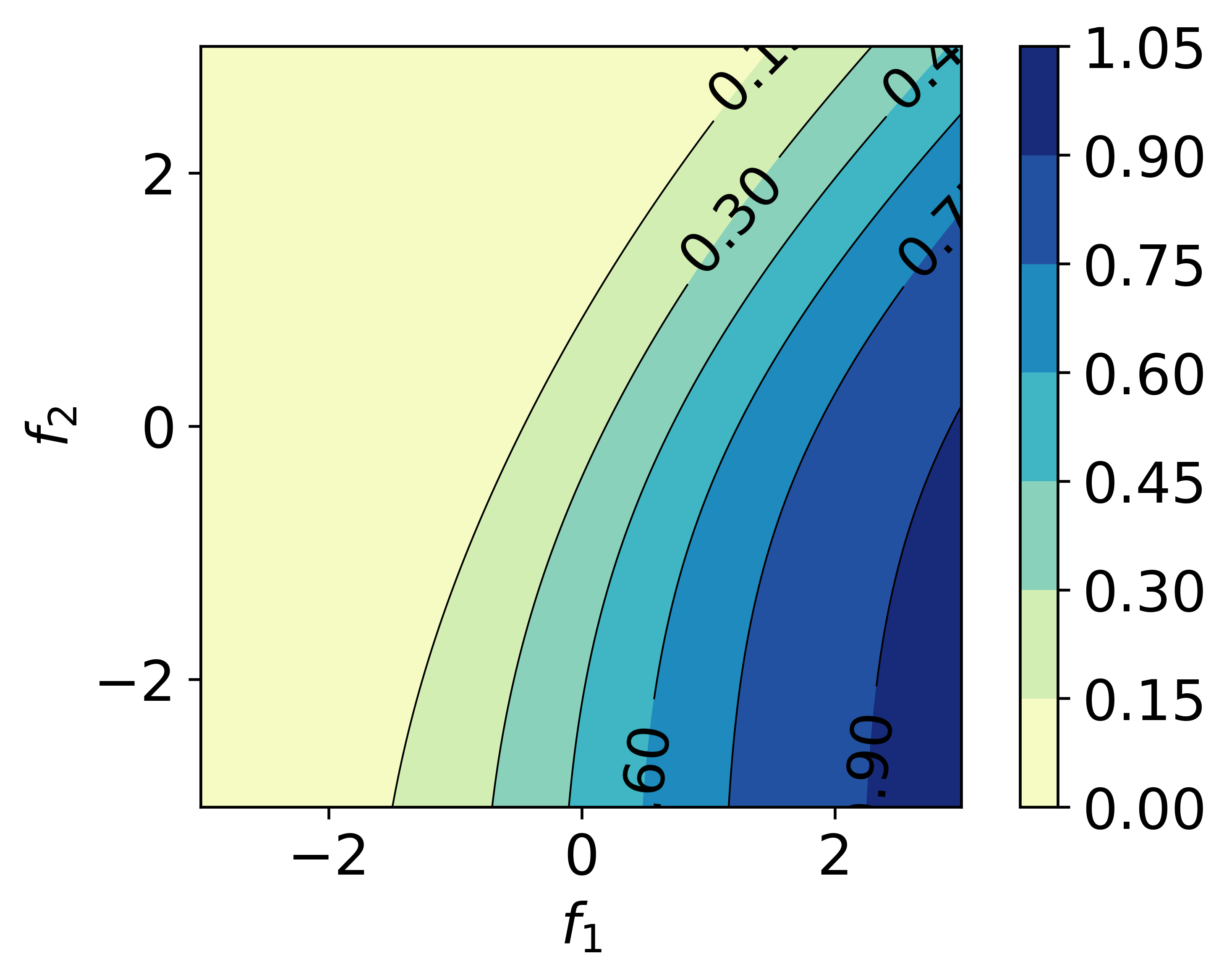}}
  \caption{Plot of $\mathcal{L}(\rvf \,|\, y=1)$, where $f_{3}$ is clamped to 0.
    The Gaussian likelihood penalizes configurations far away from
    $(f_{1}, f_{2}) = (1, -1)$. Logistic softmax is much flatter compared to
    softmax and has visibly different contours. One-vs-Each is visually similar
    to the softmax but penalizes $(f_{1}, f_{2})$ near the origin slightly
    more.}\label{fig:likelihood_plot}
\end{figure}

\begin{figure}[!ht]
  \centering
  \subfigure[Softmax]{\includegraphics[width=0.24\linewidth]{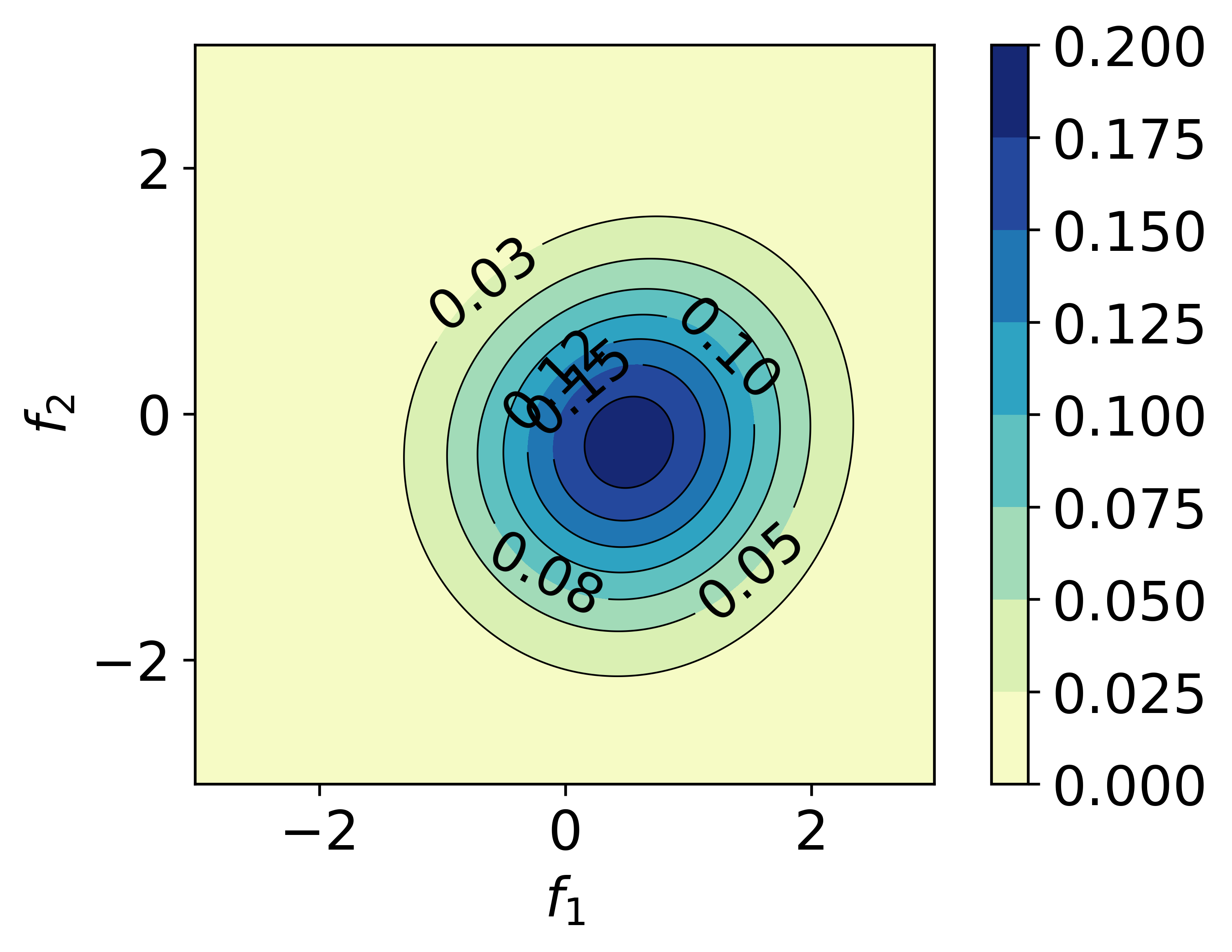}}
  \subfigure[Gaussian]{\includegraphics[width=0.24\linewidth]{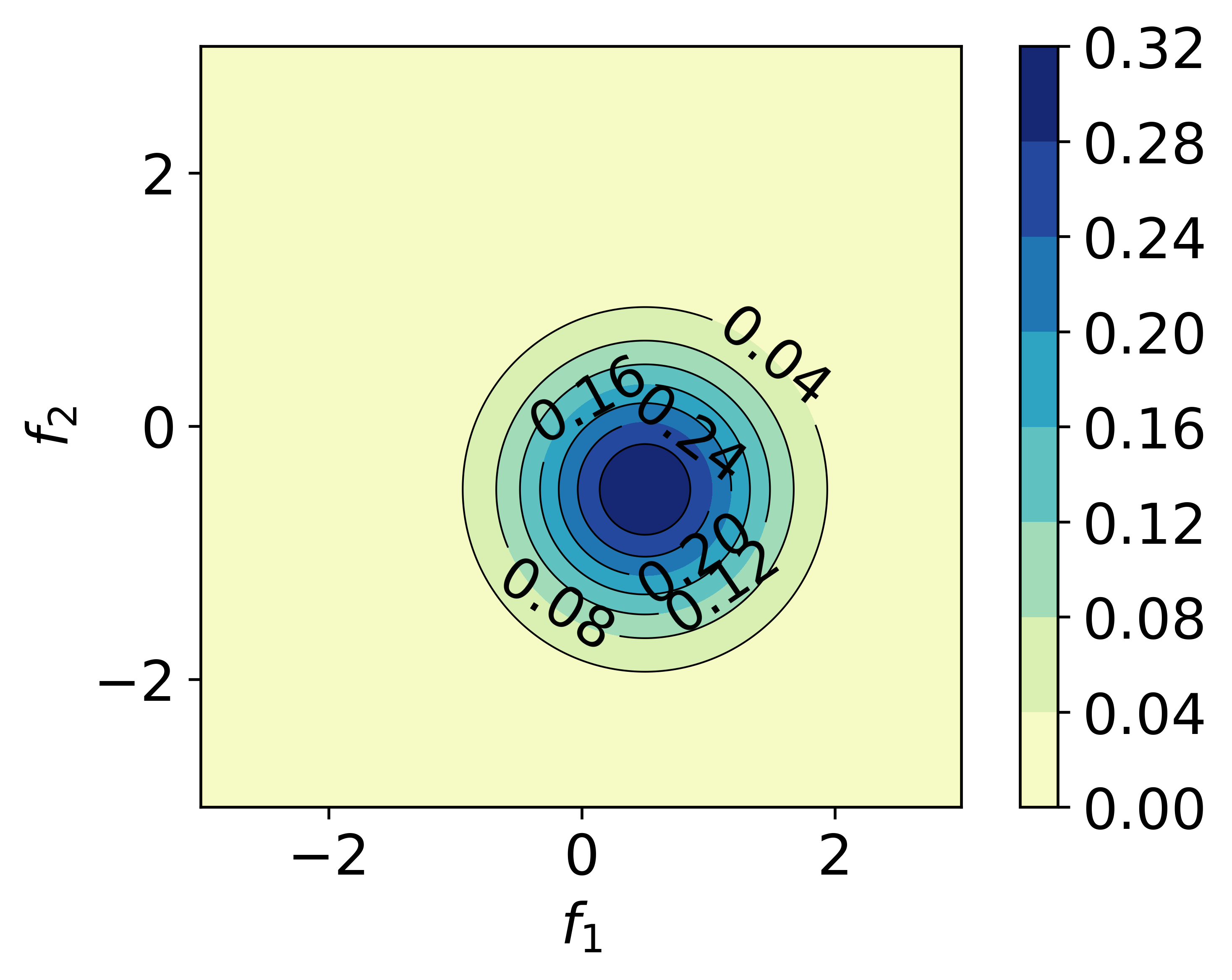}}
  \subfigure[Logistic
  Softmax]{\includegraphics[width=0.24\linewidth]{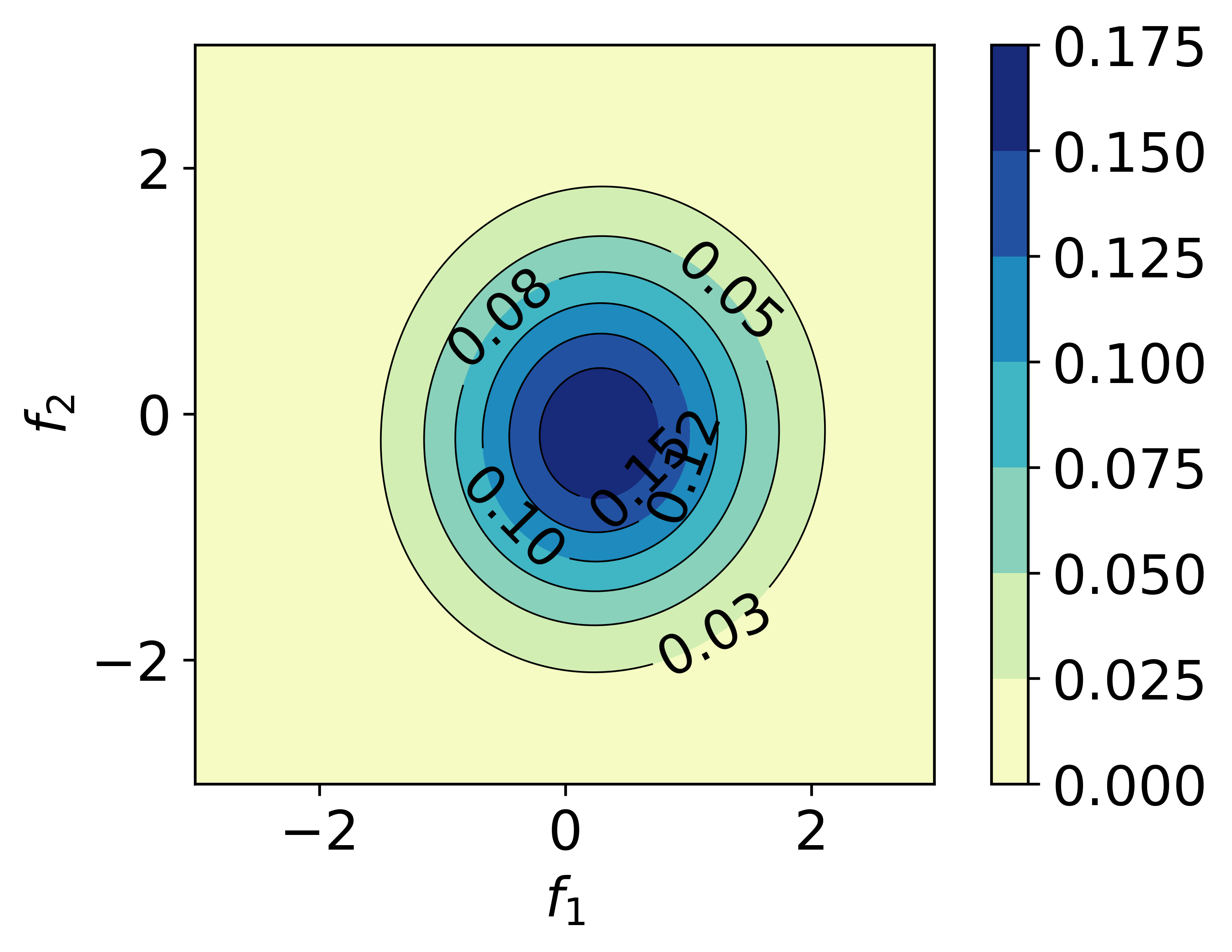}}
  \subfigure[One-vs-Each]{\includegraphics[width=0.24\linewidth]{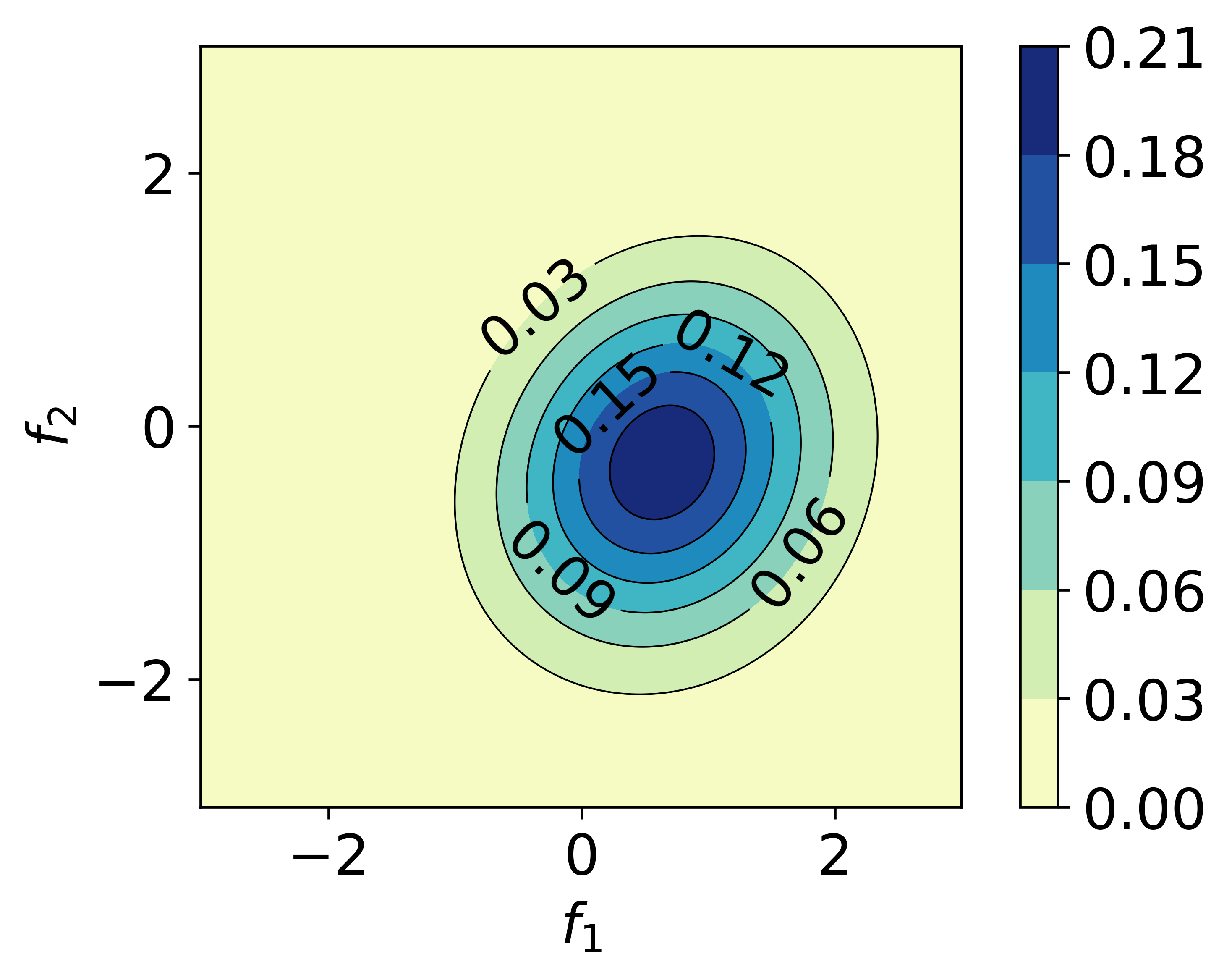}}
  \caption{Plot of posterior $p(\rvf \,|\, y=1)$, where $f_{3}$ is clamped to 0.
    The mode of each posterior distribution is similar, but each differs
    slightly in shape. Gaussian is more peaked about its mode, while logistic
    softmax is more spread out. One-vs-Each is similar to softmax, but is
    slightly more elliptical. }\label{fig:posterior_plot}
\end{figure}

In order to visualize the various likelihoods under consideration, we consider a
trivial classification task with a single observed example. We assume that there
are three classes ($C=3$) and the single example belongs to the first class
($y = 1$). We place the following prior on
$\rvf = {(f_{1}, f_{2}, f_{3})}^{\top}$:
\begin{align}
  p(\rvf) = \mathcal{N} \left( \rvf \, \biggr\rvert \, \bm{\mu} = \left[
  \begin{array}{c} 0 \\ 0 \\ 0 \end{array}
  \right], \bm{\Sigma} =\left[ \begin{array}{ccc} 1 & 0 & 0 \\ 0 & 1 & 0 \\ 0 & 0 & 0 \end{array} \right]
                                                                                    \right).
\end{align}
In other words, the prior for $f_{1}$ and $f_{2}$ is a standard normal and
$f_{3}$ is clamped at zero (for ease of visualization). The likelihoods are
plotted in Figure~\ref{fig:likelihood_plot} and the corresponding posteriors are
plotted in Figure~\ref{fig:posterior_plot}.

\subsubsection{2D Iris Experiments}

We also conducted experiments on a 2D version of the Iris dataset
\citep{fisher1936use}, which contains 150 examples across 3 classes. The first
two features of the dataset were retained (sepal length and width). We used a
zero-mean GP prior and an RBF kernel
$\displaystyle k(\rvx, \rvx') = \exp\left(-\frac{1}{2} d(\rvx, \rvx')^{2}\right)$,
where $d(\cdot, \cdot)$ is Euclidean distance. We considered training set sizes
with 1, 2, 3, 4, 5, 10, 15, 20, 25, and 30 examples per class. For each training
set size, we performed GP inference on 200 randomly generated train/test splits
and compared the predictions across Gaussian, logistic softmax, and one-vs-each
likelihoods.

\begin{figure}[!ht]
  \centering
  \subfigure[Gaussian]{\includegraphics[width=0.3\linewidth]{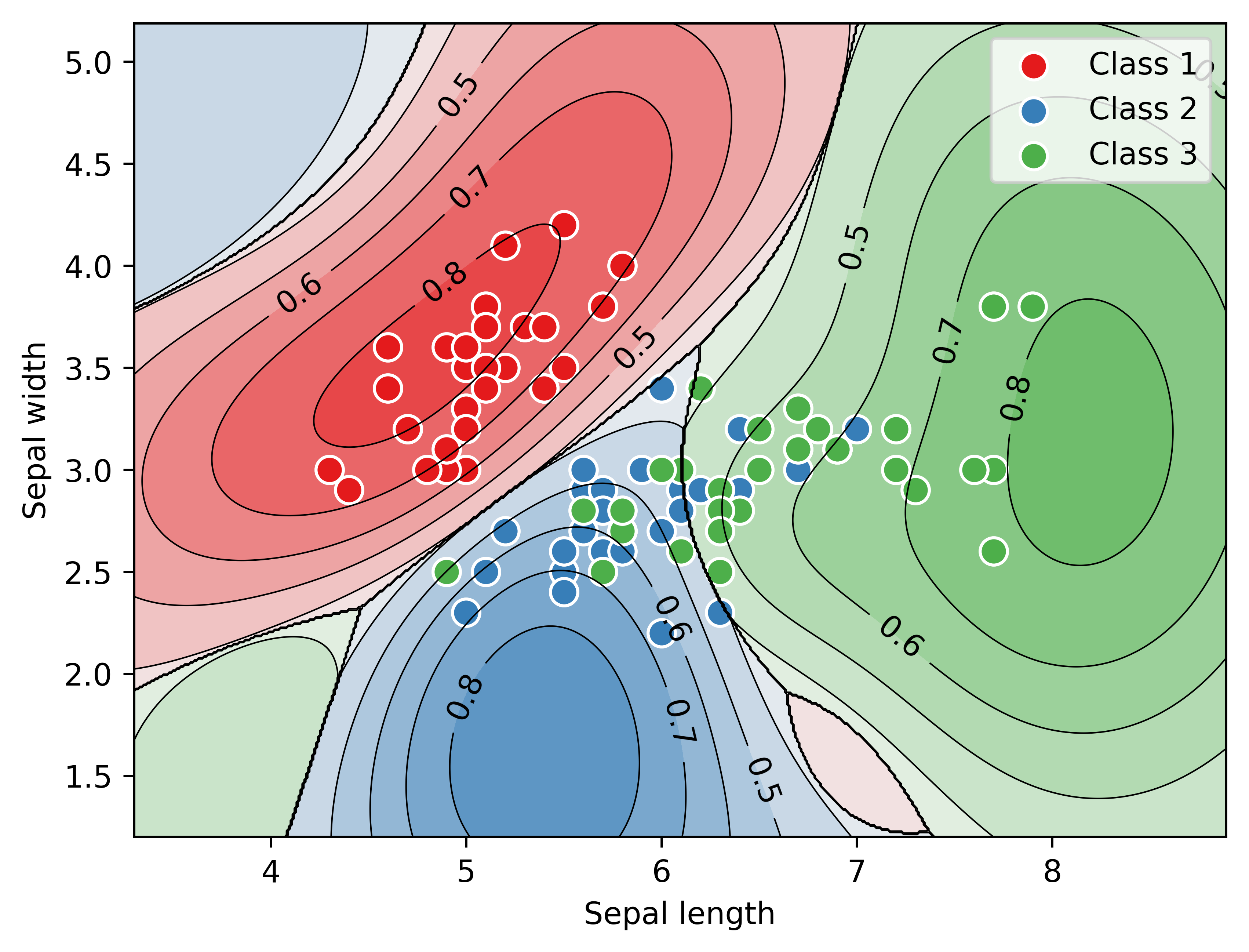}}
  \subfigure[Logistic
  Softmax]{\includegraphics[width=0.3\linewidth]{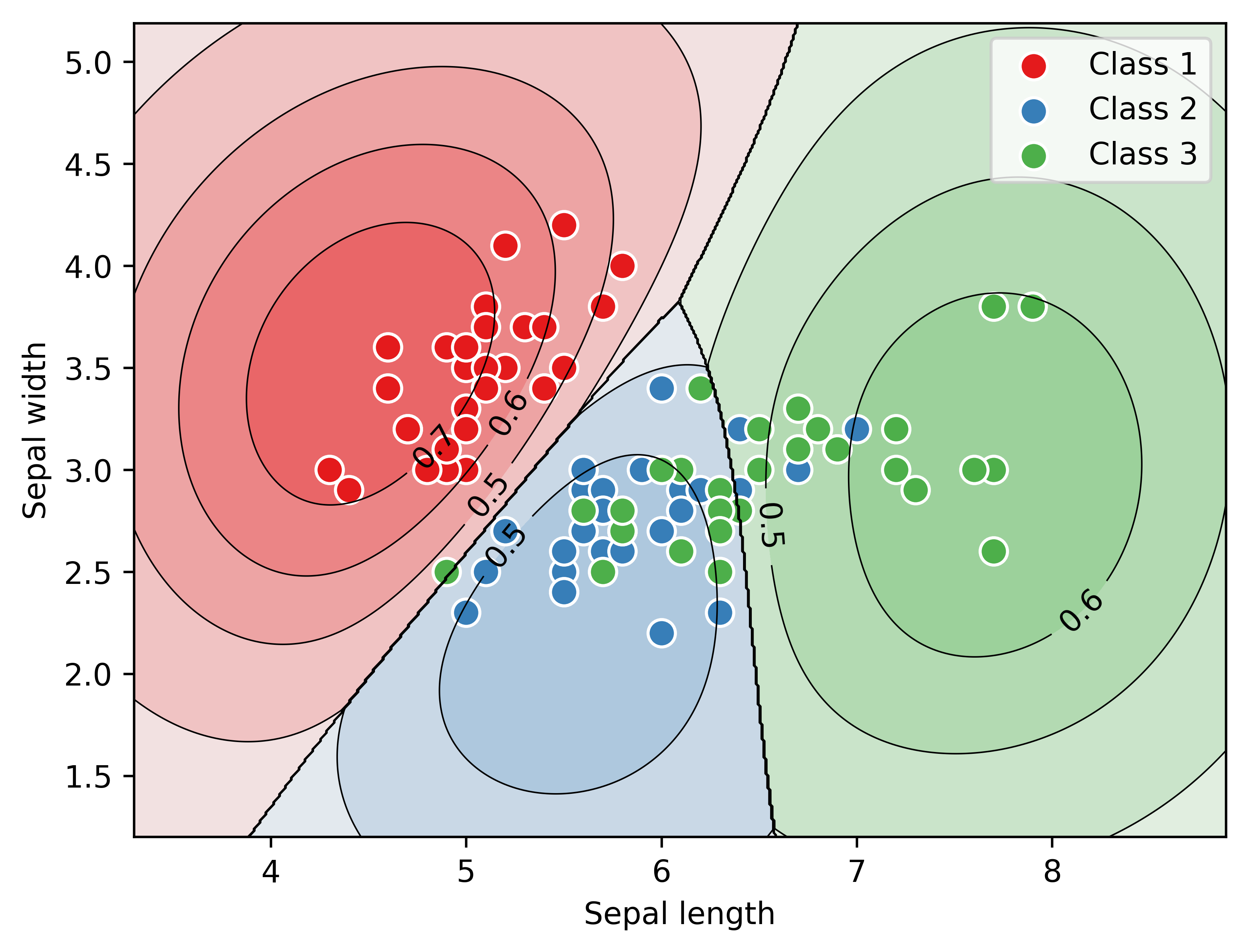}}
  \subfigure[One-vs-Each]{\includegraphics[width=0.3\linewidth]{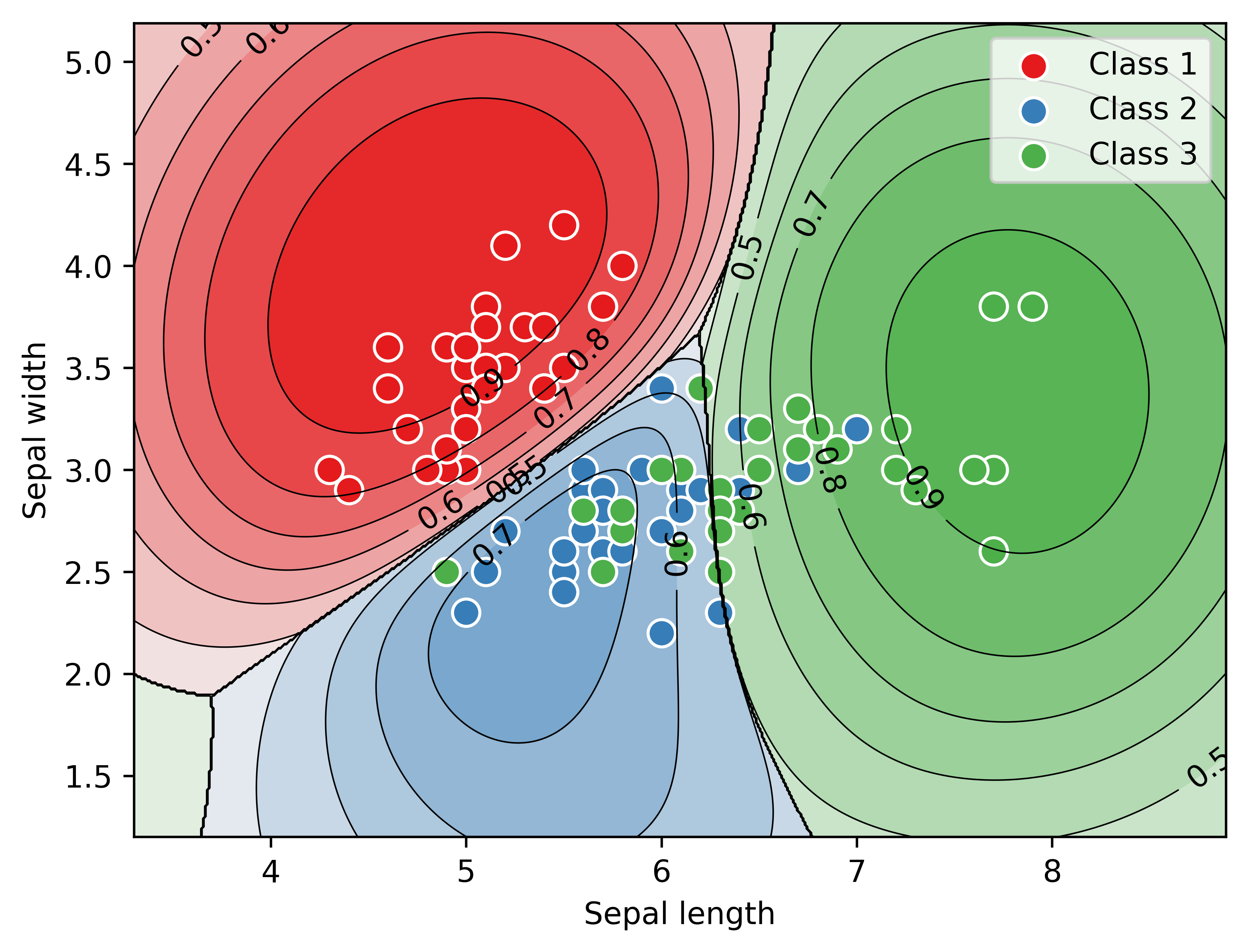}}
  \caption{Training points (colored points) and maximum predictive probability
    for various likelihoods on the Iris dataset. The Gaussian likelihood
    produces more warped decision boundaries than the others. Logistic softmax
    tends to produce lower confidence predictions, while one-vs-each produces
    larger regions of greater confidence than the others.}\label{fig:iris_proba}
\end{figure}

\begin{figure}[!ht]
  \centering
  \subfigure[Accuracy]{\includegraphics[width=0.24\linewidth]{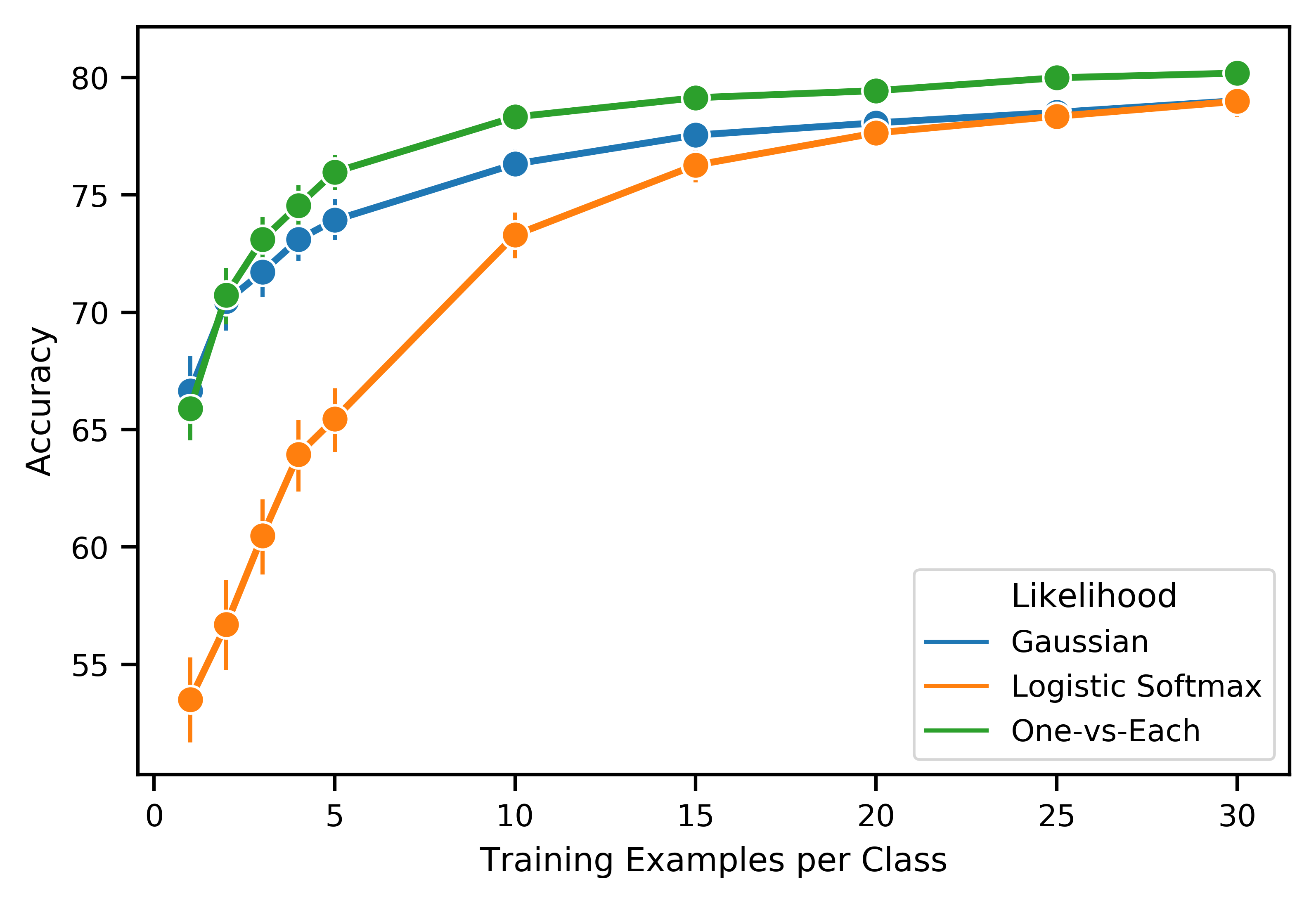}}
  \subfigure[Brier]{\includegraphics[width=0.24\linewidth]{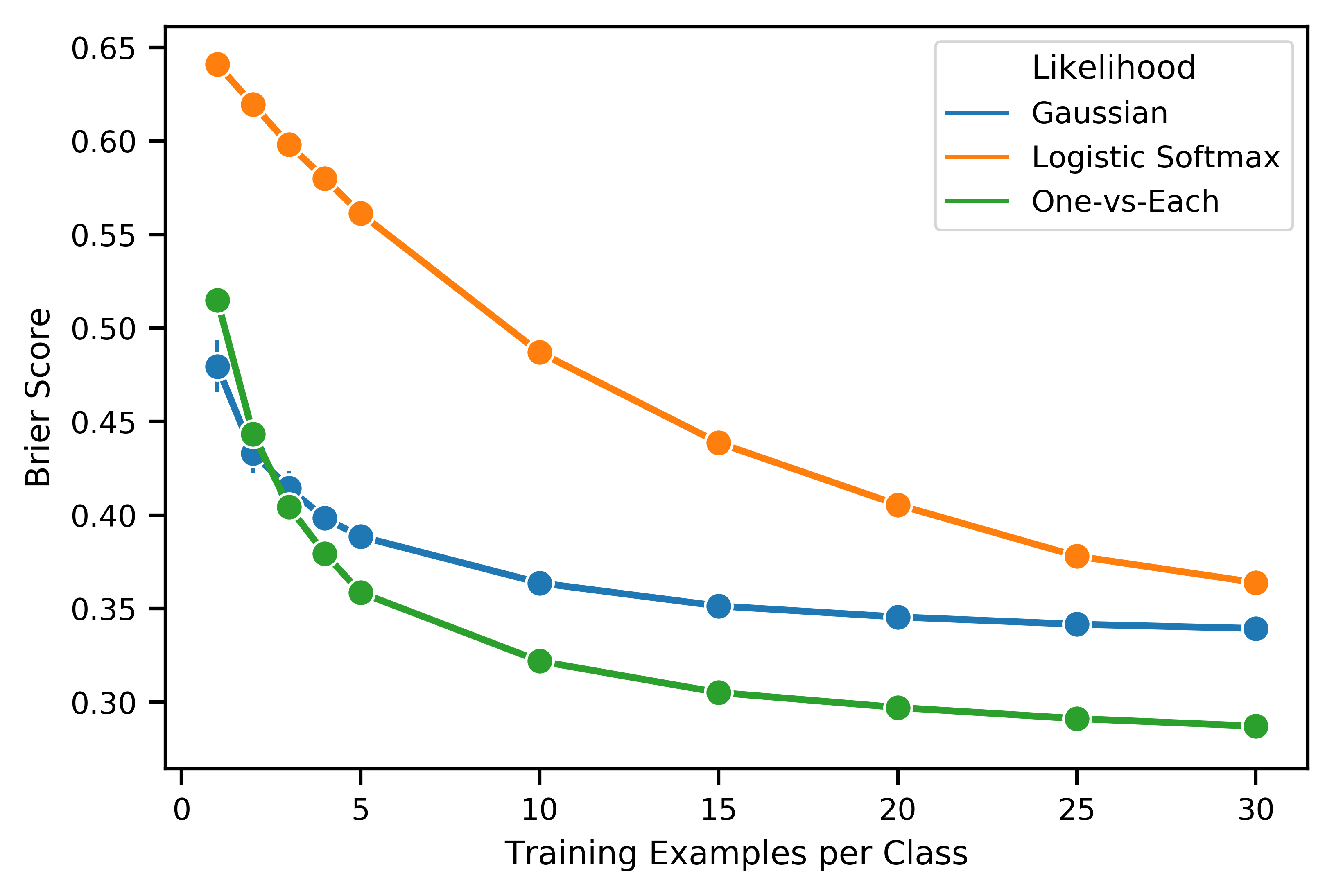}}
  \subfigure[ECE]{\includegraphics[width=0.24\linewidth]{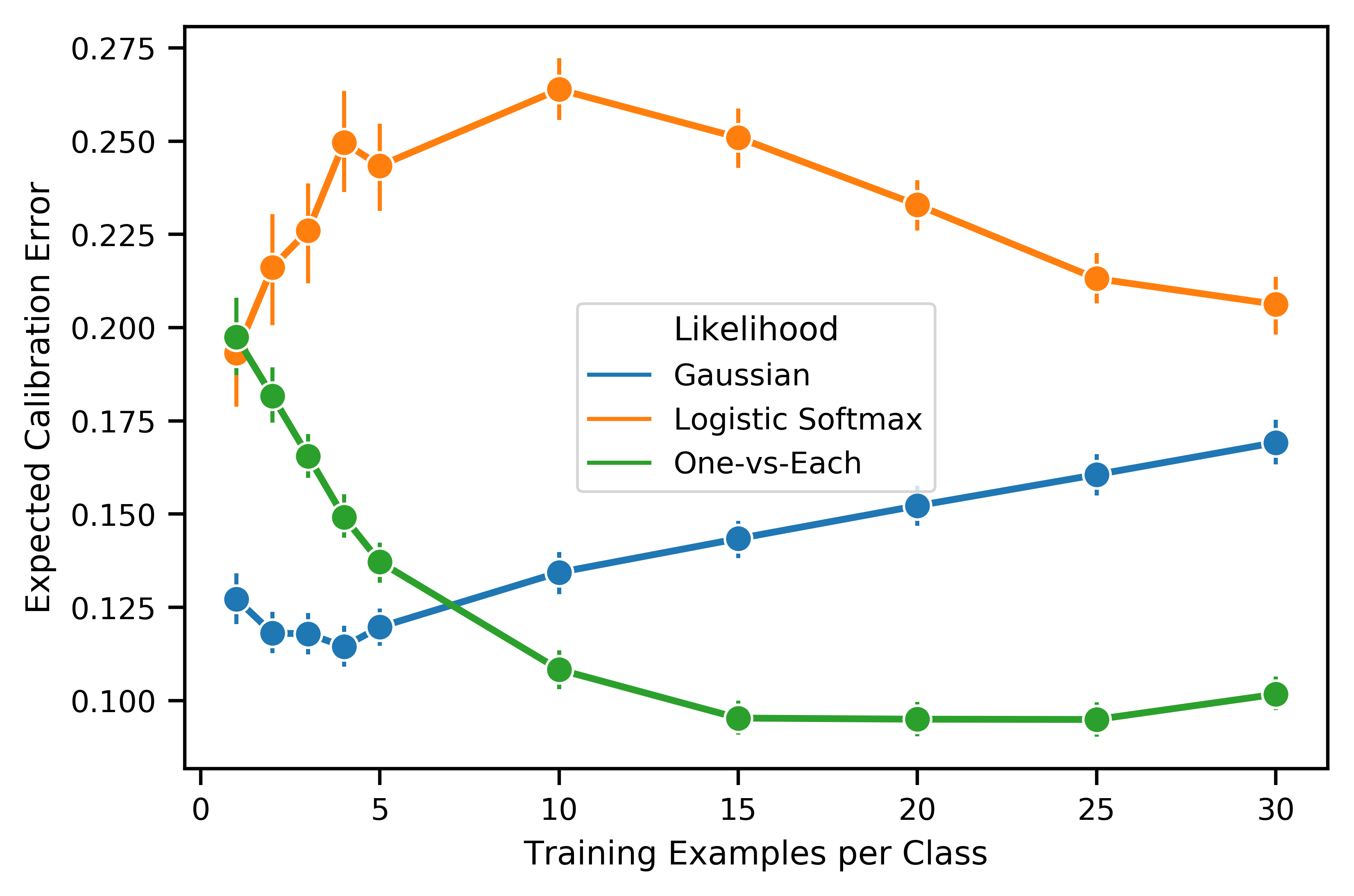}}
  \subfigure[ELBO]{\includegraphics[width=0.24\linewidth]{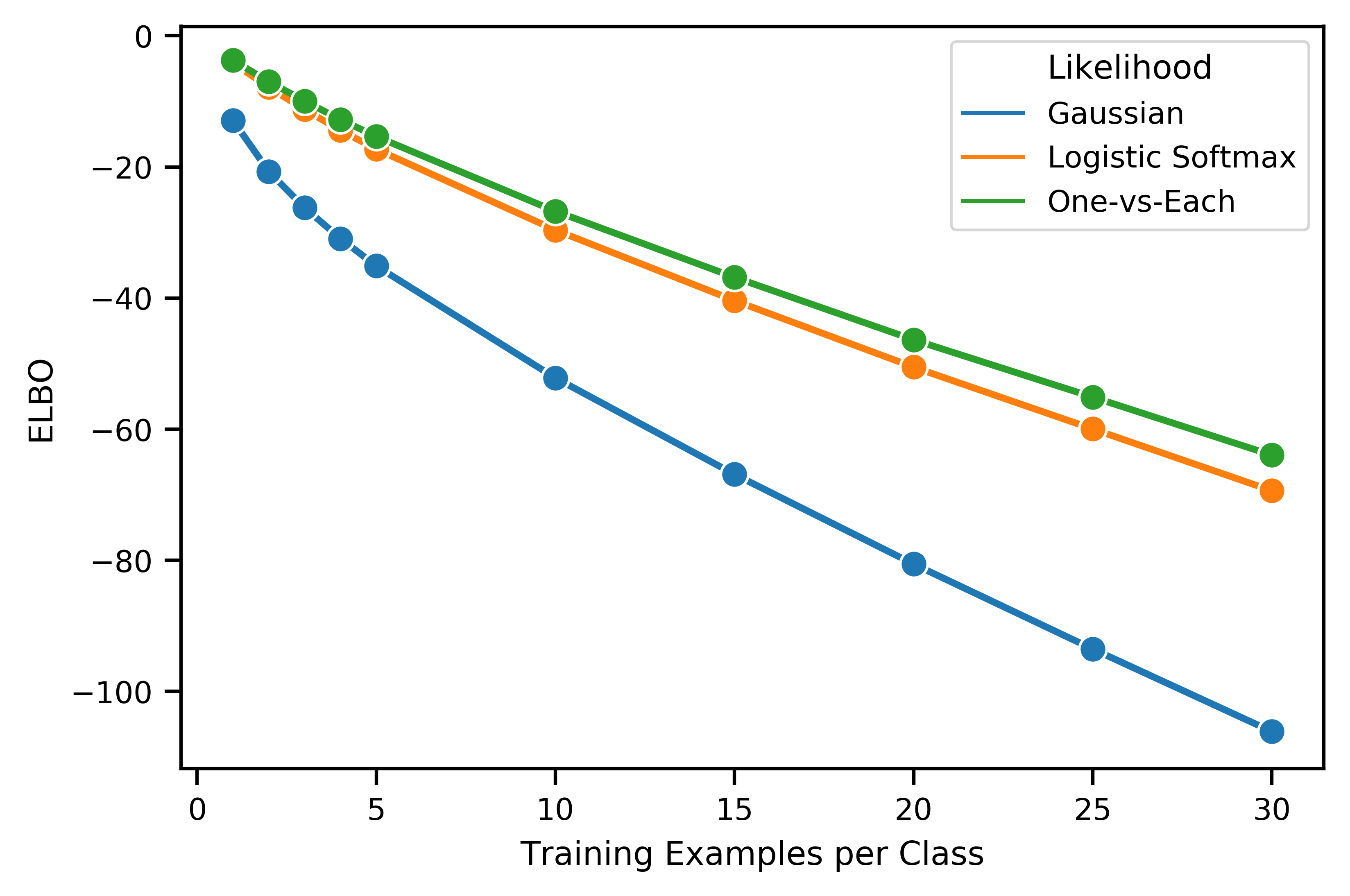}}
  \caption{Comparison across likelihoods in terms of test predictive accuracy,
    Brier score, expected calibration error (computed with 10 bins), and ELBO.
    Results are averaged over 200 randomly generated splits for each training
    set size (1, 2, 3, 4, 5, 10, 15, 20, 25, and 30 examples per class). Error
    bars indicate 95\% confidence intervals.}\label{fig:iris_sweep}
\end{figure}

Predictions at a test point $\rvx_{*}$ were made by applying the (normalized)
likelihood to the posterior predictive mean $\bar{\rvf}_{*}$. The predictive
probabilities for each likelihood is shown in Figure~\ref{fig:iris_proba} for a
randomly generated train/test split with 30 examples per class. Test predictive
accuracy, Brier score, expected calibration error, and evidence lower bound
(ELBO) results across various training set sizes are shown in
Figure~\ref{fig:iris_sweep}.

The ELBO is computed by treating each likelihood's posterior
$q(\rvf | \rmX, \rmY)$ as an approximation to the softmax posterior
$p(\rvf | \rmX, \rmY)$.
\begin{align*}
  \text{ELBO}(q) &= \mathbb{E}_q [\log  p(\mathbf{f} | \rmX)] + \mathbb{E}_q[\log p(\mathbf{Y} | \mathbf{f})] - \mathbb{E}_q [\log q(\mathbf{f} | \rmX, \rmY)] \\
                 &= \log p(\mathbf{x}) - \text{KL}( q(\mathbf{f} | \rmX, \rmY) || p(\mathbf{f} | \rmX, \rmY) ).
\end{align*}
Even though direct computation of the softmax posterior $p(\rvf | \rmX, \rvy)$
is intractable, computing the ELBO is tractable. A larger ELBO indicates a lower
KL divergence to the softmax posterior.

One-vs-Each performs well for accuracy, Brier score, and ELBO across the
training set sizes. Gaussian performs best on expected calibration error through
15 examples per class, beyond which one-vs-each is better.

\subsection{Overview of Few-shot Classification
  Experiments}\label{sec:fewshot_experiment_overview}

We now compare classification accuracy and uncertainty quantification to
representative baselines for several major approaches to few-shot
classification: fine-tuning, metric learning, gradient-based meta-learning,
Bayesian neural network, and GP-based classifiers. An overview of the baselines
we used can be found in Section~\ref{sec:baselines}.

One of our aims is to compare methods based on uncertainty quantification. We
therefore developed some new benchmark evaluations and tasks: few-shot
calibration, robustness, and out-of-episode detection. In order to empirically
compare methods, we could not simply borrow the accuracy results from other
papers, but instead needed to train each of these baselines ourselves. For all
baselines except Bayesian MAML, ABML, and Logistic Softmax GP, we ran the code
from \citep{patacchiola2020bayesian} and verified that the accuracies matched
closely to their reported results. Additional experimental details may be found
in Section~\ref{sec:experimental_details}. We have made PyTorch code for our
experiments publicly
available\footnote{\url{https://github.com/jakesnell/ove-polya-gamma-gp}}.

\subsection{Few-shot Classification Baselines}\label{sec:baselines}

The baselines we compared to are explained here in more detail.

\begin{itemize}
  \item \textbf{Feature Transfer} \citep{chen2019closer} involves first training
        an off-line classifier on the training classes and then training a new
        classification layer on the episode.
  \item \textbf{Baseline$++$} \citep{chen2019closer} is similar to Feature
        Transfer except it uses a cosine distance module prior to the softmax
        during fine-tuning.
  \item \textbf{Matching Networks} \citep{vinyals2016matching} can be viewed as
        a soft form of $k$-nearest neighbors that computes attention and sums
        over the support examples to form a predictive distribution over
        classes.
  \item \textbf{Prototypical Networks} \citep{snell2017prototypical} computes
        class means (prototypes) and forms a predictive distribution based on
        Euclidean distance to the prototypes. It can be viewed as a Gaussian
        classifier operating in an embedding space.
  \item \textbf{MAML} \citep{finn2017modelagnostic} performs one or a few steps
        of gradient descent on the support set and then makes predictions on the
        query set, backpropagating through the gradient descent procedure. For
        this baseline, we simply quote the classification accuracy reported by
        \citep{patacchiola2020bayesian}.
  \item \textbf{RelationNet} \citep{sung2018learning} rather than using a
        predefined distance metric as in Matching Networks or Prototypical
        Networks instead learns a deep distance metric as the output of a neural
        network that accepts as input the latent representation of both
        examples. It is trained to minimize squared error of output predictions.
  \item \textbf{Deep Kernel Transfer (DKT)} \citep{patacchiola2020bayesian}
        relies on least squares classification~\citep{rifkin2004defense} to
        maintain tractability of Gaussian process posterior inference. In DKT, a
        separate binary classification task is formed for each class in
        one-vs-rest fashion by treating labels in $\{-1, +1\}$ as continuous
        targets. We include the results of DKT with the cosine kernel as
        implemented by \citet{patacchiola2020bayesian}, which is parameterized
        slightly differently from the version we used in
        \eqref{eq:cosine_kernel}:
        \begin{equation}
          k^\text{cos}_\text{dkt}(\xx, \xx'; \btheta, \alpha, \nu) = \text{softplus}(\alpha) \cdot \text{softplus}(\nu) \cdot \frac{g_{\btheta}(\xx)^\top g_{\btheta}(\xx')}{\|g_{\btheta}(\xx)\| \|g_{\btheta}(\xx')\|}.
        \end{equation}
  \item \textbf{Bayesian MAML} \citep{yoon2018bayesian} relies on Stein
        Variational Gradient Descent (SVGD) \citep{liu2016stein} to get an
        approximate posterior distribution in weight-space. We compare to both
        the non-chaser version, which optimizes cross-entropy of query
        predictions, and the chaser version, which optimizes mean squared error
        between the approximate posterior on the support set and the approximate
        posterior on the merged support \& query set. The non-chaser version is
        therefore related to predictive likelihood methods and the chaser
        version is more analogous to the marginal likelihood methods. For the
        non-chaser version, we used 20 particles and 1 step of adaptation at
        both train and test time. For the chaser version, we also used 20
        particles. At train time, the chaser took 1 step and the leader 1
        additional step. At test time, we used 5 steps of adaptation. Due to the
        slow performance of this method, we followed the advice of
        \citet{yoon2018bayesian} and only performed adaptation on the final
        layer of weights, which may help explain the drop in performance
        relative to MAML. The authors released Tensorflow code for regression
        only, so we reimplemented this baseline for classification in PyTorch.
  \item \textbf{Amortized Bayesian Meta-Learning (ABML)}
        \citep{ravi2019amortized} performs a few steps of Bayes-by-backprop
        \citep{blundell2015weight} in order to infer a fully-factorized
        approximate posterior over the weights. The authors did not release code
        and so we implemented our own version of ABML in PyTorch. We found the
        weighting on the inner and outer KL divergences to be important for
        achieving good performance. We took the negative log likelihood to be
        mean cross entropy and used an inner KL weight of 0.01 and an outer KL
        weight of 0.001. These values were arrived upon by doing a small amount
        of hyperparameter tuning on the Omniglot$\rightarrow$ EMNIST dataset. We
        used $\alpha = 1.0$ and $\beta=0.01$ for the Gamma prior over the
        weights. We only applied ABML to the weights of the network; the biases
        were learned as point estimates. We used 4 steps of adaptation and took
        5 samples when computing expectations (using any more than this did not
        fit into GPU memory). We used the local reparameterization
        trick~\citep{kingma2015variational} and flipout~\citep{wen2018flipout}
        when computing expectations in order to reduce variance. In order to
        match the architecture used by \citet{ravi2019amortized}, we trained
        this baseline with 32 filters throughout the classification network.
  \item \textbf{Logistic Softmax GP} \citep{galy-fajou2020multiclass} is the
        multi-class Gaussian process classification method that relies on the
        logistic softmax likelihood. \citet{galy-fajou2020multiclass} did not
        consider few-shot, but we use the same objectives described in
        Section~\ref{sec:polya_few_shot} to adapt this method to FSC. In
        addition, we used the cosine kernel (see Section~\ref{sec:kernel_choice}
        for a description) that we found to work best with our OVE PG GPs. For
        this method, we found it important to learn a constant mean function
        (rather than a zero mean) in order to improve calibration.
\end{itemize}

\subsection{Classification on Few-shot
  Benchmarks}\label{sec:fewshot_classification_experiments}

As mentioned above, we follow the training and evaluation protocol of
\citet{patacchiola2020bayesian} for this section. We train both 1-shot and
5-shot versions of our model in four different settings: Caltech-UCSD Birds
(CUB) \citep{wah2011caltechucsd}, mini-Imagenet with the split proposed by
\citet{ravi2017optimization}, as well as two cross-domain transfer tasks:
training on mini-ImageNet and testing on CUB, and from Omniglot
\citep{lake2011one} to EMNIST \citep{cohen2017emnist}. We employ the
commonly-used Conv4 architecture with 64 channels \citep{vinyals2016matching}
for all experiments. Further experimental details and comparisons across methods
can be found in Section~\ref{sec:experimental_details}. Classification results
are shown in Table \ref{tab_classification_accuracy} and
\ref{tab_crossdomain_accuracy}. We find that our proposed P\'olya-Gamma OVE GPs
yield strong classification results, outperforming the baselines in five of the
eight scenarios.
\begin{table*}[!th]
  \caption{Average accuracy and standard deviation (percentage) on 5-way FSC.
    Baseline results (through DKT) are from \citet{patacchiola2020bayesian}.
    Evaluation is performed on 3,000 randomly generated test episodes. Standard
    deviation for our approach is computed by averaging over 5 batches of 600
    episodes with different random seeds. The best results are highlighted in
    bold.}
  \centering
  \begin{tabular}{lcccc}
    \hline
    \textbf{} & \multicolumn{2}{c}{\textbf{CUB}} & \multicolumn{2}{c}{\textbf{mini-ImageNet}} \\
    \small{\textbf{Method}} & \textbf{1-shot} & \textbf{5-shot} & \textbf{1-shot} & \textbf{5-shot}\\
    \hline
    \small{\textbf{Feature Transfer}}               & 46.19 $\pm$ \small{0.64} & 68.40 $\pm$ \small{0.79} & 39.51 $\pm$ \small{0.23} & 60.51 $\pm$ \small{0.55}\\
    \small{\textbf{Baseline$++$}} & 61.75 $\pm$ \small{0.95} & 78.51 $\pm$ \small{0.59} & 47.15 $\pm$ \small{0.49} & 66.18 $\pm$ \small{0.18}\\
    \small{\textbf{MatchingNet}} & 60.19 $\pm$ \small{1.02} & 75.11 $\pm$ \small{0.35} & 48.25 $\pm$ \small{0.65} & 62.71 $\pm$ \small{0.44} \\
    \small{\textbf{ProtoNet}} & 52.52 $\pm$ \small{1.90} & 75.93 $\pm$ \small{0.46} &44.19 $\pm$ \small{1.30} & 64.07 $\pm$ \small{0.65} \\
    \small{\textbf{RelationNet}} & 62.52 $\pm$ \small{0.34} & 78.22 $\pm$ \small{0.07}  & 48.76 $\pm$ \small{0.17} & 64.20 $\pm$ \small{0.28}\\
    \small{\textbf{MAML}}  & 56.11 $\pm$ \small{0.69} & 74.84 $\pm$ \small{0.62}  &45.39 $\pm$ \small{0.49} & 61.58 $\pm$ \small{0.53} \\
    \small{\textbf{DKT + Cosine}} & 63.37 $\pm$ \small{0.19} & 77.73 $\pm$ \small{0.26} & 48.64 $\pm$ \small{0.45} & 62.85 $\pm$ \small{0.37} \\
    \small{\textbf{Bayesian MAML}} & 55.93 $\pm$ \small{0.71} & 72.87 $\pm$ \small{0.26} & 44.46 $\pm$ \small{0.30} & 62.60 $\pm$ 0.25 \\
    \small{\textbf{Bayesian MAML (Chaser)}} & 53.93 $\pm$ \small{0.72} & 71.16  $\pm$ \small{0.32} & 43.74 $\pm$ \small{0.46} & 59.23 $\pm$ \small{0.34} \\
    \small{\textbf{ABML}} & 49.57 $\pm$ \small{0.42} & 68.94 $\pm$ \small{0.16} & 37.65 $\pm$ \small{0.22} & 56.08 $\pm$ \small{0.29} \\

    \small{\textbf{Logistic Softmax GP + Cosine (ML)}} & 60.23 $\pm$ \small{0.54} & 74.58 $\pm$ \small{0.25} & 46.75 $\pm$ \small{0.20} & 59.93 $\pm$ \small{0.31} \\
    \small{\textbf{Logistic Softmax GP + Cosine (PL)}} & 60.07 $\pm$ \small{0.29} & 78.14 $\pm$ \small{0.07} & 47.05 $\pm$ \small{0.20} & 66.01 $\pm$ \small{0.25} \\
    \hline

    \small{\textbf{OVE PG GP + Cosine (ML)}} [ours] & \textbf{63.98 $\pm$ \small{0.43}} & 77.44 $\pm$ \small{0.18} & \textbf{50.02 $\pm$ \small{0.35}} & 64.58 $\pm$ \small{0.31} \\
    \small{\textbf{OVE PG GP + Cosine (PL)}} [ours] & 60.11 $\pm$ \small{0.26} & \textbf{79.07 $\pm$ \small{0.05}} & 48.00 $\pm$ \small{0.24} & \textbf{67.14 $\pm$ \small{0.23}} \\

    \hline
  \end{tabular}
  \label{tab_classification_accuracy}
\end{table*}
\begin{table*}[!ht]
  \caption{Average accuracy and standard deviation (percentage) on 5-way
    cross-domain FSC, with the same experimental setup as in Table
    \ref{tab_classification_accuracy}. Baseline results (through DKT) are from
    \citep{patacchiola2020bayesian}.}
  \centering
  \begin{tabular}{lcccc}
    \hline
    \textbf{} & \multicolumn{2}{c}{\textbf{Omniglot}$\rightarrow$\textbf{EMNIST}} & \multicolumn{2}{c}{\textbf{mini-ImageNet}$\rightarrow$\textbf{CUB}} \\
    \small{\textbf{Method}} & \textbf{1-shot}& \textbf{5-shot} & \textbf{1-shot} & \textbf{5-shot} \\
    \hline
    \small{\textbf{Feature Transfer}} & 64.22 $\pm$ \small{1.24} & 86.10 $\pm$ \small{0.84} & 32.77 $\pm$ \small{0.35} & 50.34 $\pm$ \small{0.27}\\
    \small{\textbf{Baseline$++$}} & 56.84 $\pm$ \small{0.91} & 80.01 $\pm$ \small{0.92} & 39.19 $\pm$ \small{0.12} & \textbf{57.31 $\pm$ \small{0.11}}\\
    \small{\textbf{MatchingNet}}  & 75.01 $\pm$ \small{2.09} & 87.41 $\pm$ \small{1.79} & 36.98 $\pm$ \small{0.06} & 50.72 $\pm$ \small{0.36} \\
    \small{\textbf{ProtoNet}} & 72.04 $\pm$ \small{0.82} & 87.22 $\pm$ \small{1.01} & 33.27 $\pm$ \small{1.09} & 52.16 $\pm$ \small{0.17} \\
    \small{\textbf{RelationNet}} & 75.62 $\pm$ \small{1.00} & 87.84 $\pm$ \small{0.27}  & 37.13 $\pm$ \small{0.20} & 51.76 $\pm$ \small{1.48}\\
    \small{\textbf{MAML}} & 72.68 $\pm$ \small{1.85} & 83.54 $\pm$ \small{1.79}  & 34.01 $\pm$ \small{1.25} &48.83 $\pm$ \small{0.62} \\
    \small{\textbf{DKT + Cosine}} & 73.06 $\pm$ \small{2.36} & \textbf{88.10 $\pm$ \small{0.78}} & \textbf{40.22 $\pm$ \small{0.54}} & 55.65 $\pm$ \small{0.05}\\
    \small{\textbf{Bayesian MAML}} & 63.94 $\pm$ \small{0.47} & 65.26 $\pm$ \small{0.30} & 33.52 $\pm$ \small{0.36} & 51.35 $\pm$ \small{0.16} \\
    \small{\textbf{Bayesian MAML (Chaser)}} & 55.04 $\pm$ \small{0.34} & 54.19 $\pm$ \small{0.32} & 36.22 $\pm$ \small{0.50} & 51.53 $\pm$ \small{0.43}  \\
    \small{\textbf{ABML}} & 76.37 $\pm$ \small{0.29} & 87.96 $\pm$ \small{0.28} & 29.35 $\pm$ \small{0.26} & 45.74 $\pm$ \small{0.33} \\

    \small{\textbf{Logistic Softmax GP + Cosine (ML)}} & 62.91 $\pm$ \small{0.49} & 83.80 $\pm$ \small{0.13} & 36.41 $\pm$ 0.18 & 50.33 $\pm$ 0.13 \\
    \small{\textbf{Logistic Softmax GP + Cosine (PL)}} & 70.70 $\pm$ \small{0.36} & 86.59 $\pm$ 0.15 & 36.73 $\pm$ \small{0.26} & 56.70 $\pm$ \small{0.31}  \\
    \hline

    \small{\textbf{OVE PG GP + Cosine (ML)}} [ours] & 68.43  $\pm$ \small{0.67} & 86.22 $\pm$ \small{0.20} & 39.66 $\pm$ \small{0.18} & 55.71 $\pm$ \small{0.31} \\
    \small{\textbf{OVE PG GP + Cosine (PL)}} [ours] & \textbf{77.00 $\pm$ \small{0.50}} & 87.52 $\pm$ \small{0.19} & 37.49 $\pm$ \small{0.11} & 57.23 $\pm$ \small{0.31} \\
    \hline
  \end{tabular}
  \label{tab_crossdomain_accuracy}
\end{table*}
\subsection{Uncertainty Quantification through
  Calibration} \label{sec:calibration_results}

We next turn to uncertainty quantification, an important concern for few-shot
classifiers. When used in safety-critical applications such as medical
diagnosis, it is important for a machine learning system to defer when there is
not enough evidence to make a decision. Even in non-critical applications,
precise uncertainty quantification helps practitioners in the few-shot setting
determine when a class has an adequate amount of labeled data or when more
labels are required, and can facilitate active learning.

\begin{figure*}[ht]
  \begin{center}
    \centerline{\includegraphics[width=\textwidth]{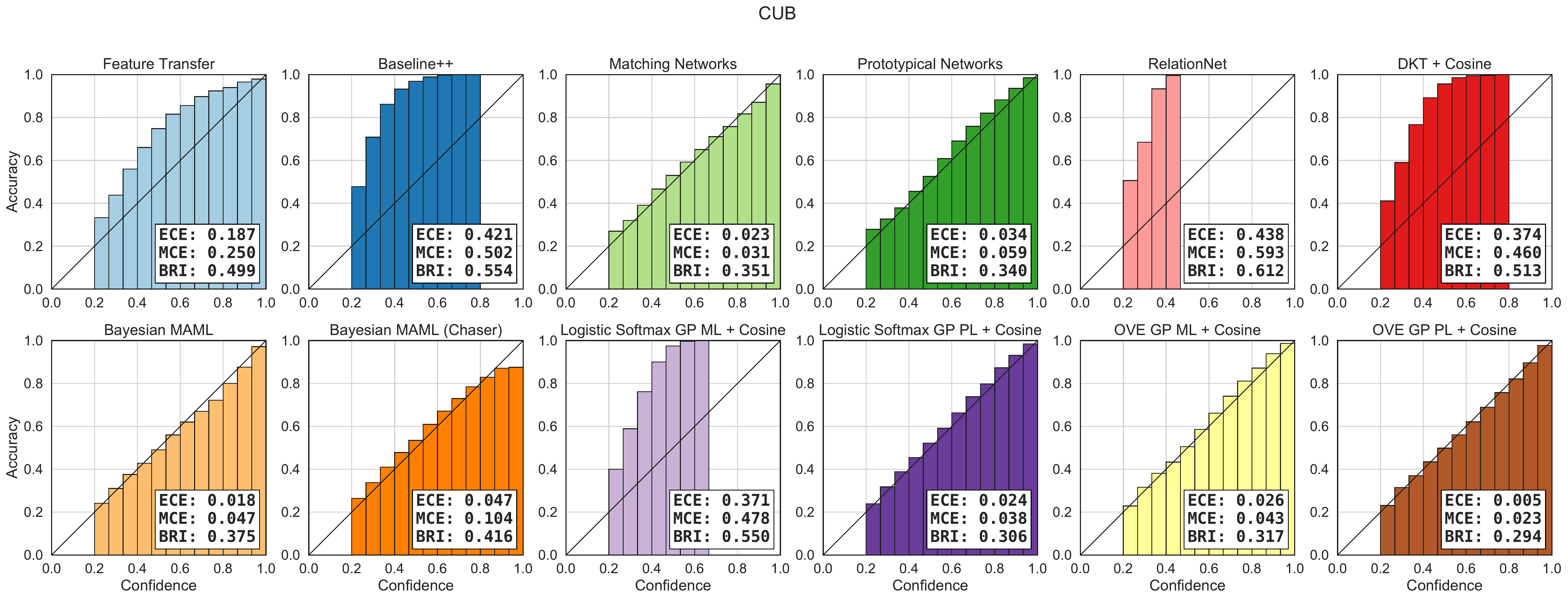}}
    \vskip -0.05in
    \caption{Reliability diagrams, expected calibration error (ECE), maximum
      calibration error (MCE), and Brier Score (BRI) for 5-shot 5-way tasks on
      CUB (additional calibration results can be found in
      Appendix~\ref{sec:additional_calibration}). Metrics are computed on 3,000
      random tasks from the test set. The last two plots are our proposed
      method.}
    \label{fig:calibration}
  \end{center}
  \vskip -0.2in
\end{figure*}

We chose several commonly used metrics for calibration. Expected calibration
error (ECE) \citep{guo2017calibration} measures the expected binned difference
between confidence and accuracy. Maximum calibration error (MCE) is similar to
ECE but measures maximum difference instead of expected difference. Brier score
(BRI) \citep{brier1950verification} is a proper scoring rule computed as the
squared error between the output probabilities and the one-hot label. For a
recent perspective on metrics for uncertainty evaluation, please refer to
\citet{ovadia2019can}. The results for representative approaches on 5-shot,
5-way CUB can be found in Figure~\ref{fig:calibration}. Our OVE PG GPs
are the best calibrated overall across the metrics.

\subsection{Robustness to Input Noise} 
\begin{figure*}[ht]
  \begin{center}
    \centerline{\includegraphics[width=\textwidth]{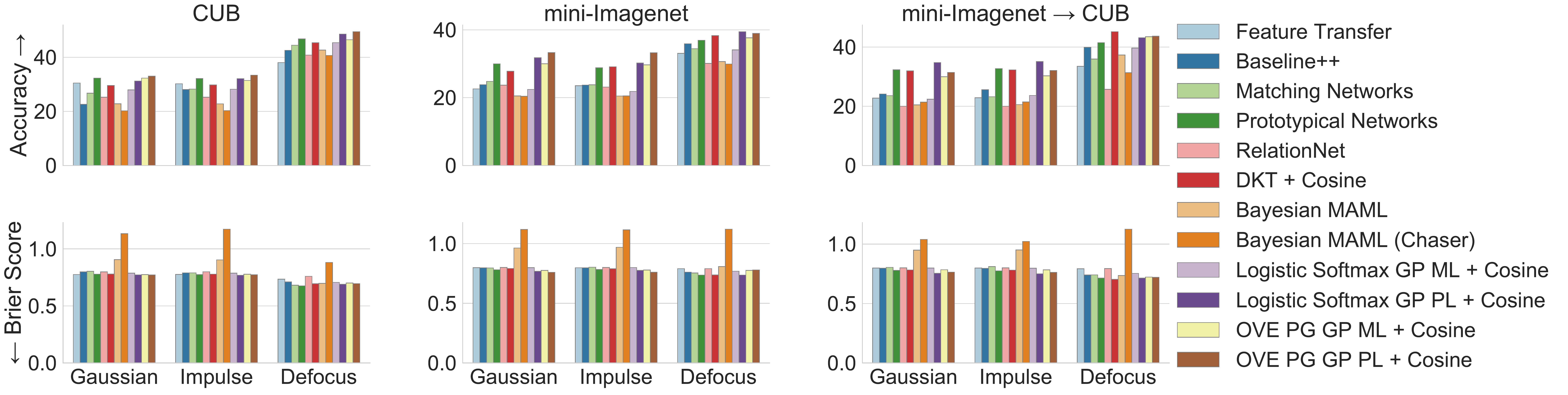}}
    \vskip -0.05in
    \caption{Accuracy ($\uparrow$) and Brier Score ($\downarrow$) when
      corrupting both support and query with Gaussian noise on 5-way 5-shot
      tasks. Additional results may be found in
      Appendix~\ref{sec:additional_robustness}.}
    \label{fig:corruptions}
  \end{center}
\end{figure*}

Input examples for novel classes in FSC
may have been collected under conditions that do not match those observed at
training time. For example, labeled support images in a medical diagnosis
application may come from a different hospital than the training set. To mimic a
simplified version of this scenario, we investigate robustness to input noise.
We used the Imagecorruptions package \citep{michaelis2019benchmarking} to apply
Gaussian noise, impulse noise, and defocus blur to both the support set and
query sets of episodes at test time and evaluated both accuracy and calibration.
We used corruption severity of 5 (severe) and evaluated across 1,000 randomly
generated tasks on the three datasets involving natural images. The robustness
results for Gaussian noise are shown in Figure \ref{fig:corruptions}. Full
quantitative results tables for each noise type may be found in
Section~\ref{sec:additional_robustness}. We find that in general Bayesian
approaches tend to be robust due to their ability to marginalize over hypotheses
consistent with the support labels. Our approach is one of the top performing
methods across all settings.

\subsection{Out-of-Episode Detection}\label{sec:fewshot_ooe}
\begin{figure*}[ht]
  \begin{center}
    \centerline{\includegraphics[width=\textwidth]{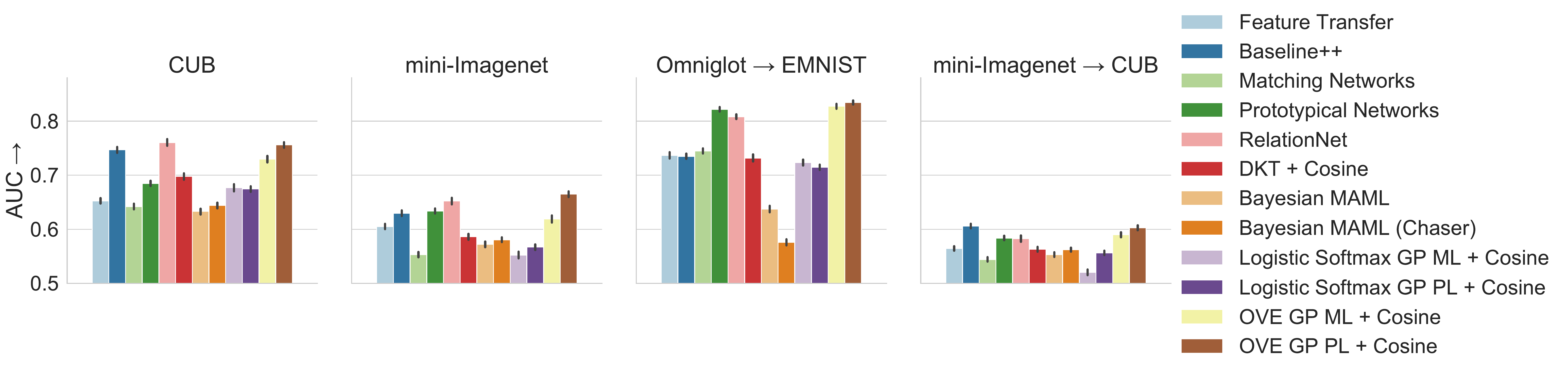}}
    \vskip -0.15in
    \caption{Average AUROC ($\uparrow$) for out-of-episode detection. The AUC is
      computed separately for each episode and averaged across 1,000 episodes.
      Bars indicate a 95\% bootstrapped confidence interval.}
    \label{fig:ooe_detection}
  \end{center}
\end{figure*}

Finally, we measure performance on out-of-episode detection, another application
in which uncertainty quantification is important. In this experiment, we used
5-way, 5-shot support sets at test time but incorporated out-of-episode examples
into the query set. Each episode had 150 query examples: 15 from each of 5
randomly chosen in-episode classes and 15 from each of 5 randomly chosen
out-of-episode classes. We then computed the AUROC of binary outlier detection
using the negative of the maximum logit as the score. Intuitively, if none of
the support classes assign a high logit to the example, it can be classified as
an outlier. The results are shown in Figure \ref{fig:ooe_detection}. Our
approach generally performs the best across the datasets.

\section{Discussion}

In Section~\ref{sec:likelihood_comparison}, we compared OVE to the softmax,
Gaussian, and logistic softmax likelihoods. We observed that OVE tends to
exhibit high output confidences (Figure~\ref{fig:logit_simulation} and
\ref{fig:iris_proba}c). The view of OVE as a composite likelihood described in
Section~\ref{sec:ove_composite} may help explain this phenomenon, since
composite likelihoods treat each component event as independent even when they
are not. In the context of OVE, each pairwise observation is treated as
additional new evidence in favor of the class with the largest function value.
Despite this, we observed in Figures~\ref{fig:likelihood_plot} and
\ref{fig:posterior_plot} that the OVE likelihood is visually quite similar to
the softmax. More importantly, we saw that it yielded higher accuracy and
competitive calibration results on the Iris dataset experiments in
Figure~\ref{fig:iris_sweep}.

This strong accuracy and calibration carried over to the few-shot experiments in
Sections~\ref{sec:fewshot_classification_experiments}-\ref{sec:fewshot_ooe}. In
terms of baselines, we observed that the fine-tuning approaches can be very good
in terms of accuracy (Baseline$++$ in particular), but do not produce reliable
estimates of uncertainty. Methods relying on Gaussian likelihoods (RelationNet
and DKT) tend to also exhibit poor uncertainty quantification. We hypothesize
this is due to the ill-suited nature of applying Gaussian likelihoods to the
fundamentally discrete task of classification. We also found that optimizing for
predictive cross-entropy generally improves classification accuracy and to some
extent can remedy calibration issues of marginal likelihood-based methods. The
OVE likelihood appears to be better suited to classification than the Logistic
Softmax likelihood, as can be seen by comparing the accuracy and calibration
results of the ML versions of these models.

Overall, our proposed OVE PG GP classifier demonstrates strong performance
across a wide range of scenarios. Future work involves investigating further the
relationship between one-vs-each and composite likelihoods. In particular,
previous works have proposed weighting component likelihoods in order to
encourage the composite posterior to more closely match the exact
posterior~\citep{ribatet2012bayesian,stoehr2015calibration}. Studying the extent
to which such weighting is necessary or useful for classification would be
valuable.

\section{Conclusion}

In this work, we have proposed a Bayesian few-shot classification approach based
on Gaussian processes. Our method replaces the ordinary softmax likelihood with
a one-vs-each pairwise composite likelihood and applies P\'olya-Gamma
augmentation to perform inference. This allows us to model class logits directly
as function values and efficiently marginalize over uncertainty in each few-shot
episode. Modeling functions directly enables our approach to avoid the
dependence on model size that posterior inference in weight-space based models
inherently have. Our approach compares favorably to baseline FSC methods under a
variety of dataset and shot configurations, including dataset transfer. We also
demonstrate strong uncertainty quantification, robustness to input noise, and
out-of-episode detection. We believe that Bayesian modeling is a powerful tool
for handling uncertainty and hope that our work will lead to broader adoption of
efficient Bayesian inference in the few-shot scenario.

\section*{Acknowledgments}
We would like to thank Ryan Adams, Ethan Fetaya, Mike Mozer, Eleni
Triantafillou, Kuan-Chieh Wang, Max Welling, and Amos Storkey for helpful
discussions. JS also thanks SK T-Brain for supporting him on an internship that
led to precursors of some ideas in this paper. Resources used in preparing this
research were provided, in part, by the Province of Ontario, the Government of
Canada through CIFAR, and companies sponsoring the Vector Institute
(\url{https://www.vectorinstitute.ai/partners}). This project is supported by
NSERC and the Intelligence Advanced Research Projects Activity (IARPA) via
Department of Interior/Interior Business Center (DoI/IBC) contract number
D16PC00003. The U.S. Government is authorized to reproduce and distribute
reprints for Governmental purposes notwithstanding any copyright annotation
thereon. Disclaimer: The views and conclusions contained herein are those of the
authors and should not be interpreted as necessarily representing the official
policies or endorsements, either expressed or implied, of IARPA, DoI/IBC, or the
U.S. Government.

\bibliography{references}

\begin{thebibliography}{61}
\providecommand{\natexlab}[1]{#1}
\providecommand{\url}[1]{\texttt{#1}}
\expandafter\ifx\csname urlstyle\endcsname\relax
  \providecommand{\doi}[1]{doi: #1}\else
  \providecommand{\doi}{doi: \begingroup \urlstyle{rm}\Url}\fi

\bibitem[Albert and Chib(1993)]{albert1993bayesian}
James~H. Albert and Siddhartha Chib.
\newblock Bayesian {{Analysis}} of {{Binary}} and {{Polychotomous Response
  Data}}.
\newblock \emph{Journal of the American Statistical Association}, 88\penalty0
  (422):\penalty0 669--679, 1993.

\bibitem[Besag(1975)]{besag1975statistical}
Julian Besag.
\newblock Statistical analysis of non-lattice data.
\newblock \emph{Journal of the Royal Statistical Society: Series D (The
  Statistician)}, 24\penalty0 (3):\penalty0 179--195, 1975.

\bibitem[Blundell et~al.(2015)Blundell, Cornebise, Kavukcuoglu, and
  Wierstra]{blundell2015weight}
Charles Blundell, Julien Cornebise, Koray Kavukcuoglu, and Daan Wierstra.
\newblock Weight {{Uncertainty}} in {{Neural Networks}}.
\newblock In \emph{International {{Conference}} on {{Machine Learning}}}, 2015.

\bibitem[Brier(1950)]{brier1950verification}
Glenn~W. Brier.
\newblock Verification of forecasts expressed in terms of probability.
\newblock \emph{Monthly weather review}, 78\penalty0 (1):\penalty0 1--3, 1950.

\bibitem[Chen et~al.(2019)Chen, Liu, Kira, Wang, and Huang]{chen2019closer}
Wei-Yu Chen, Yen-Cheng Liu, Zsolt Kira, Yu-Chiang~Frank Wang, and Jia-Bin
  Huang.
\newblock A closer look at few-shot classification.
\newblock In \emph{International Conference on Learning Representations}, 2019.

\bibitem[Cohen et~al.(2017)Cohen, Afshar, Tapson, and {van
  Schaik}]{cohen2017emnist}
Gregory Cohen, Saeed Afshar, Jonathan Tapson, and Andr{\'e} {van Schaik}.
\newblock {{EMNIST}}: {{Extending MNIST}} to handwritten letters.
\newblock In \emph{2017 International Joint Conference on Neural Networks
  ({{IJCNN}})}, 2017.

\bibitem[Douc et~al.(2014)Douc, Moulines, and Stoffer]{douc2014nonlinear}
Randal Douc, Eric Moulines, and David Stoffer.
\newblock \emph{Nonlinear Time Series: {{Theory}}, Methods and Applications
  with {{R}} Examples}.
\newblock {CRC press}, 2014.

\bibitem[Doucet(2010)]{doucet2010note}
Arnaud Doucet.
\newblock A {{Note}} on {{Efficient Conditional Simulation}} of {{Gaussian
  Distributions}}.
\newblock 2010.
\newblock URL
  \url{https://www.cs.ubc.ca/~arnaud/doucet_simulationconditionalgaussian.pdf}.

\bibitem[Finn et~al.(2017)Finn, Abbeel, and Levine]{finn2017modelagnostic}
Chelsea Finn, Pieter Abbeel, and Sergey Levine.
\newblock Model-agnostic meta-learning for fast adaptation of deep networks.
\newblock In Doina Precup and Yee~Whye Teh, editors, \emph{Proceedings of the
  34th International Conference on Machine Learning}, volume~70 of
  \emph{Proceedings of Machine Learning Research}, {International Convention
  Centre, Sydney, Australia}, 2017. {PMLR}.

\bibitem[Finn et~al.(2018)Finn, Xu, and Levine]{finn2018probabilistic}
Chelsea Finn, Kelvin Xu, and Sergey Levine.
\newblock Probabilistic model-agnostic meta-learning.
\newblock In S.~Bengio, H.~Wallach, H.~Larochelle, K.~Grauman,
  N.~{Cesa-Bianchi}, and R.~Garnett, editors, \emph{Advances in Neural
  Information Processing Systems}, volume~31. {Curran Associates, Inc.}, 2018.

\bibitem[Fisher(1936)]{fisher1936use}
R.~A. Fisher.
\newblock The {{Use}} of {{Multiple Measurements}} in {{Taxonomic Problems}}.
\newblock \emph{Annals of Eugenics}, 7\penalty0 (2):\penalty0 179--188, 1936.

\bibitem[{Galy-Fajou} et~al.(2020){Galy-Fajou}, Wenzel, Donner, and
  Opper]{galy-fajou2020multiclass}
Th{\'e}o {Galy-Fajou}, Florian Wenzel, Christian Donner, and Manfred Opper.
\newblock Multi-class gaussian process classification made conjugate:
  {{Efficient}} inference via data augmentation.
\newblock In Ryan~P. Adams and Vibhav Gogate, editors, \emph{Proceedings of the
  35th Uncertainty in Artificial Intelligence Conference}, volume 115 of
  \emph{Proceedings of Machine Learning Research}, {Tel Aviv, Israel}, 2020.
  {PMLR}.

\bibitem[Girolami and Rogers(2006)]{girolami2006variational}
Mark Girolami and Simon Rogers.
\newblock Variational {{Bayesian Multinomial Probit Regression}} with
  {{Gaussian Process Priors}}.
\newblock \emph{Neural Computation}, 18\penalty0 (8):\penalty0 1790--1817,
  2006.

\bibitem[Gordon et~al.(2019)Gordon, Bronskill, Bauer, Nowozin, and
  Turner]{gordon2019metalearning}
Jonathan Gordon, John Bronskill, Matthias Bauer, Sebastian Nowozin, and Richard
  Turner.
\newblock Meta-learning probabilistic inference for prediction.
\newblock In \emph{International Conference on Learning Representations}, 2019.

\bibitem[Grant et~al.(2018)Grant, Finn, Levine, Darrell, and
  Griffiths]{grant2018recasting}
Erin Grant, Chelsea Finn, Sergey Levine, Trevor Darrell, and Thomas Griffiths.
\newblock Recasting gradient-based meta-learning as hierarchical bayes.
\newblock In \emph{International Conference on Learning Representations}, 2018.

\bibitem[Guo et~al.(2017)Guo, Pleiss, Sun, and Weinberger]{guo2017calibration}
Chuan Guo, Geoff Pleiss, Yu~Sun, and Kilian~Q. Weinberger.
\newblock On calibration of modern neural networks.
\newblock In Doina Precup and Yee~Whye Teh, editors, \emph{Proceedings of the
  34th International Conference on Machine Learning}, volume~70 of
  \emph{Proceedings of Machine Learning Research}, {International Convention
  Centre, Sydney, Australia}, 2017. {PMLR}.

\bibitem[{Hernandez-Lobato} and
  {Hernandez-Lobato}(2016)]{hernandez-lobato2016scalable}
Daniel {Hernandez-Lobato} and Jose~Miguel {Hernandez-Lobato}.
\newblock Scalable gaussian process classification via expectation propagation.
\newblock In Arthur Gretton and Christian~C. Robert, editors, \emph{Proceedings
  of the 19th International Conference on Artificial Intelligence and
  Statistics}, volume~51 of \emph{Proceedings of Machine Learning Research},
  {Cadiz, Spain}, 2016. {PMLR}.

\bibitem[Hilliard et~al.(2018)Hilliard, Phillips, Howland, Yankov, Corley, and
  Hodas]{hilliard2018fewshot}
Nathan Hilliard, Lawrence Phillips, Scott Howland, Art{\"e}m Yankov,
  Courtney~D. Corley, and Nathan~O. Hodas.
\newblock Few-{{Shot Learning}} with {{Metric}}-{{Agnostic Conditional
  Embeddings}}.
\newblock \emph{arXiv:1802.04376 [cs, stat]}, 2018.

\bibitem[Hoffman and Ribak(1991)]{hoffman1991constrained}
Yehuda Hoffman and Erez Ribak.
\newblock Constrained realizations of {{Gaussian}} fields-{{A}} simple
  algorithm.
\newblock \emph{The Astrophysical Journal}, 380:\penalty0 L5--L8, 1991.

\bibitem[Kim and Ghahramani(2006)]{kim2006bayesian}
Hyun-Chul Kim and Zoubin Ghahramani.
\newblock Bayesian {{Gaussian}} process classification with the {{EM}}-{{EP}}
  algorithm.
\newblock \emph{IEEE Transactions on Pattern Analysis and Machine
  Intelligence}, 28\penalty0 (12):\penalty0 1948--1959, 2006.

\bibitem[Kingma and Ba(2015)]{kingma2015adam}
Diederik~P. Kingma and Jimmy Ba.
\newblock Adam: {{A Method}} for {{Stochastic Optimization}}.
\newblock In \emph{International {{Conference}} on {{Learning
  Representations}}}, 2015.

\bibitem[Kingma et~al.(2015)Kingma, Salimans, and
  Welling]{kingma2015variational}
Diederik~P. Kingma, Tim Salimans, and Max Welling.
\newblock Variational dropout and the local reparameterization trick.
\newblock In C.~Cortes, N.~Lawrence, D.~Lee, M.~Sugiyama, and R.~Garnett,
  editors, \emph{Advances in Neural Information Processing Systems}, volume~28.
  {Curran Associates, Inc.}, 2015.

\bibitem[Koch(2015)]{koch2015siamese}
Gregory Koch.
\newblock \emph{Siamese {{Neural Networks}} for {{One}}-{{Shot Image
  Recognition}}}.
\newblock Master's {{Thesis}}, University of Toronto, 2015.

\bibitem[Lake et~al.(2011)Lake, Salakhutdinov, Gross, and
  Tenenbaum]{lake2011one}
Brenden Lake, Ruslan Salakhutdinov, Jason Gross, and Joshua Tenenbaum.
\newblock One shot learning of simple visual concepts.
\newblock In \emph{Proceedings of the Annual Meeting of the Cognitive Science
  Society}, volume~33, 2011.

\bibitem[Linderman et~al.(2015)Linderman, Johnson, and
  Adams]{linderman2015dependent}
Scott Linderman, Matthew Johnson, and Ryan~P. Adams.
\newblock Dependent {{Multinomial Models Made Easy}}: {{Stick}}-{{Breaking}}
  with the {{P\'olya}}-gamma {{Augmentation}}.
\newblock In \emph{Advances in {{Neural Information Processing Systems}}},
  2015.

\bibitem[Lindsay(1988)]{lindsay1988composite}
Bruce~G. Lindsay.
\newblock Composite {{Likelihood Methods}}.
\newblock \emph{Contemporary Mathematics}, 80:\penalty0 221--239, 1988.

\bibitem[Liu and Wang(2016)]{liu2016stein}
Qiang Liu and Dilin Wang.
\newblock Stein variational gradient descent: {{A}} general purpose bayesian
  inference algorithm.
\newblock In D.~Lee, M.~Sugiyama, U.~Luxburg, I.~Guyon, and R.~Garnett,
  editors, \emph{Advances in Neural Information Processing Systems}, volume~29.
  {Curran Associates, Inc.}, 2016.

\bibitem[Matthews et~al.(2016)Matthews, Hensman, Turner, and
  Ghahramani]{matthews2016sparse}
Alexander G. de~G. Matthews, James Hensman, Richard Turner, and Zoubin
  Ghahramani.
\newblock On sparse variational methods and the kullback-leibler divergence
  between stochastic processes.
\newblock In Arthur Gretton and Christian~C. Robert, editors, \emph{Proceedings
  of the 19th International Conference on Artificial Intelligence and
  Statistics}, volume~51 of \emph{Proceedings of Machine Learning Research},
  {Cadiz, Spain}, 2016. {PMLR}.

\bibitem[Michaelis et~al.(2019)Michaelis, Mitzkus, Geirhos, Rusak, Bringmann,
  Ecker, Bethge, and Brendel]{michaelis2019benchmarking}
Claudio Michaelis, Benjamin Mitzkus, Robert Geirhos, Evgenia Rusak, Oliver
  Bringmann, Alexander~S. Ecker, Matthias Bethge, and Wieland Brendel.
\newblock Benchmarking {{Robustness}} in {{Object Detection}}: {{Autonomous
  Driving}} when {{Winter}} is {{Coming}}.
\newblock In \emph{{{NeurIPS}} 2019 {{Machine Learning}} for {{Autonomous
  Driving Workshop}}}, 2019.

\bibitem[Miller(2019)]{miller2019asymptotic}
Jeffrey~W. Miller.
\newblock Asymptotic normality, concentration, and coverage of generalized
  posteriors.
\newblock \emph{arXiv:1907.09611 [math, stat]}, 2019.

\bibitem[Minka(2001)]{minka2001family}
Thomas~Peter Minka.
\newblock \emph{A Family of Algorithms for Approximate {{Bayesian}} Inference}.
\newblock PhD thesis, Massachusetts Institute of Technology, 2001.

\bibitem[Ovadia et~al.(2019)Ovadia, Fertig, Ren, Nado, Sculley, Nowozin,
  Dillon, Lakshminarayanan, and Snoek]{ovadia2019can}
Yaniv Ovadia, Emily Fertig, Jie Ren, Zachary Nado, D.~Sculley, Sebastian
  Nowozin, Joshua Dillon, Balaji Lakshminarayanan, and Jasper Snoek.
\newblock Can you trust your models uncertainty? {{Evaluating}} predictive
  uncertainty under dataset shift.
\newblock In H.~Wallach, H.~Larochelle, A.~Beygelzimer, F.~{dAlch{\'e}-Buc},
  E.~Fox, and R.~Garnett, editors, \emph{Advances in Neural Information
  Processing Systems}, volume~32. {Curran Associates, Inc.}, 2019.

\bibitem[Patacchiola et~al.(2020)Patacchiola, Turner, Crowley, O'Boyle, and
  Storkey]{patacchiola2020bayesian}
Massimiliano Patacchiola, Jack Turner, Elliot~J. Crowley, Michael O'Boyle, and
  Amos Storkey.
\newblock Bayesian {{Meta}}-{{Learning}} for the {{Few}}-{{Shot Setting}} via
  {{Deep Kernels}}.
\newblock In \emph{Advances in Neural Information Processing Systems}, 2020.

\bibitem[Pauli et~al.(2011)Pauli, Racugno, and Ven]{pauli2011bayesian}
Francesco Pauli, Walter Racugno, and Laura Ven.
\newblock Bayesian {{Composite Marginal Likelihoods}}.
\newblock \emph{Statistica Sinica}, 2011.

\bibitem[Perez et~al.(2018)Perez, Strub, {de Vries}, Dumoulin, and
  Courville]{perez2018film}
Ethan Perez, Florian Strub, Harm {de Vries}, Vincent Dumoulin, and Aaron
  Courville.
\newblock {{FiLM}}: {{Visual Reasoning}} with a {{General Conditioning Layer}}.
\newblock \emph{Proceedings of the AAAI Conference on Artificial Intelligence},
  32\penalty0 (1), 2018.

\bibitem[Polson et~al.(2013)Polson, Scott, and Windle]{polson2013bayesian}
Nicholas~G. Polson, James~G. Scott, and Jesse Windle.
\newblock Bayesian {{Inference}} for {{Logistic Models Using
  P\'olya}}\textendash{{Gamma Latent Variables}}.
\newblock \emph{Journal of the American Statistical Association}, 108\penalty0
  (504):\penalty0 1339--1349, 2013.

\bibitem[Prabhu(2019)]{prabhu2019fewshot}
Viraj~Uday Prabhu.
\newblock \emph{Few-Shot {{Learning For Dermatological Disease Diagnosis}}}.
\newblock Master's {{Thesis}}, Georgia Institute of Technology, 2019.

\bibitem[Rasmussen and Williams(2006)]{rasmussen2006gaussian}
Carl~Edward Rasmussen and Christopher K.~I. Williams.
\newblock \emph{Gaussian Processes for Machine Learning}.
\newblock Adaptive Computation and Machine Learning. {MIT Press}, {Cambridge,
  Mass}, 2006.
\newblock ISBN 978-0-262-18253-9.

\bibitem[Ravi and Beatson(2019)]{ravi2019amortized}
Sachin Ravi and Alex Beatson.
\newblock Amortized {{Bayesian Meta}}-learning.
\newblock In \emph{International {{Conference}} on {{Learning
  Representations}}}, 2019.

\bibitem[Ravi and Larochelle(2017)]{ravi2017optimization}
Sachin Ravi and Hugo Larochelle.
\newblock Optimization as a model for few-shot learning.
\newblock In \emph{5th International Conference on Learning Representations},
  2017.

\bibitem[Requeima et~al.(2019)Requeima, Gordon, Bronskill, Nowozin, and
  Turner]{requeima2019fast}
James Requeima, Jonathan Gordon, John Bronskill, Sebastian Nowozin, and
  Richard~E Turner.
\newblock Fast and flexible multi-task classification using conditional neural
  adaptive processes.
\newblock In H.~Wallach, H.~Larochelle, A.~Beygelzimer, F.~{dAlch{\'e}-Buc},
  E.~Fox, and R.~Garnett, editors, \emph{Advances in Neural Information
  Processing Systems}, volume~32. {Curran Associates, Inc.}, 2019.

\bibitem[Ribatet et~al.(2012)Ribatet, Cooley, and Davison]{ribatet2012bayesian}
Mathieu Ribatet, Daniel Cooley, and Anthony~C. Davison.
\newblock Bayesian {{Inference From Composite Likelihoods}}, {{With An
  Application To Spatial Extremes}}.
\newblock \emph{Statistica Sinica}, 2012.

\bibitem[Rifkin and Klautau(2004)]{rifkin2004defense}
Ryan Rifkin and Aldebaro Klautau.
\newblock In defense of one-vs-all classification.
\newblock \emph{Journal of machine learning research}, 5\penalty0
  (Jan):\penalty0 101--141, 2004.

\bibitem[Rusu et~al.(2019)Rusu, Rao, Sygnowski, Vinyals, Pascanu, Osindero, and
  Hadsell]{rusu2019metalearning}
Andrei~A. Rusu, Dushyant Rao, Jakub Sygnowski, Oriol Vinyals, Razvan Pascanu,
  Simon Osindero, and Raia Hadsell.
\newblock Meta-learning with latent embedding optimization.
\newblock In \emph{International Conference on Learning Representations}, 2019.

\bibitem[Snell et~al.(2017)Snell, Swersky, and Zemel]{snell2017prototypical}
Jake Snell, Kevin Swersky, and Richard Zemel.
\newblock Prototypical networks for few-shot learning.
\newblock In I.~Guyon, U.~V. Luxburg, S.~Bengio, H.~Wallach, R.~Fergus,
  S.~Vishwanathan, and R.~Garnett, editors, \emph{Advances in Neural
  Information Processing Systems}, volume~30. {Curran Associates, Inc.}, 2017.

\bibitem[Stoehr and Friel(2015)]{stoehr2015calibration}
Julien Stoehr and Nial Friel.
\newblock Calibration of conditional composite likelihood for {{Bayesian}}
  inference on {{Gibbs}} random fields.
\newblock In \emph{International {{Conference}} on {{Artificial Intelligence}}
  and {{Statistics}}}, 2015.

\bibitem[Sun et~al.(2019)Sun, Zhang, Shi, and Grosse]{sun2019functional}
Shengyang Sun, Guodong Zhang, Jiaxin Shi, and Roger Grosse.
\newblock Functional {{Variational Bayesian Neural Networks}}.
\newblock In \emph{International {{Conference}} on {{Learning
  Representations}}}, 2019.

\bibitem[Sung et~al.(2018)Sung, Yang, Zhang, Xiang, Torr, and
  Hospedales]{sung2018learning}
Flood Sung, Yongxin Yang, Li~Zhang, Tao Xiang, Philip~H.S. Torr, and Timothy~M.
  Hospedales.
\newblock Learning to compare: {{Relation}} network for few-shot learning.
\newblock In \emph{Proceedings of the {{IEEE}} Conference on Computer Vision
  and Pattern Recognition ({{CVPR}})}, 2018.

\bibitem[Titsias(2009)]{titsias2009variational}
Michalis Titsias.
\newblock Variational learning of inducing variables in sparse gaussian
  processes.
\newblock In David {van Dyk} and Max Welling, editors, \emph{Proceedings of the
  Twelth International Conference on Artificial Intelligence and Statistics},
  volume~5 of \emph{Proceedings of Machine Learning Research}, {Hilton
  Clearwater Beach Resort, Clearwater Beach, Florida USA}, 2009. {PMLR}.

\bibitem[Titsias(2016)]{titsias2016onevseach}
Michalis~K. Titsias.
\newblock One-vs-{{Each Approximation}} to {{Softmax}} for {{Scalable
  Estimation}} of {{Probabilities}}.
\newblock In \emph{Advances in {{Neural Information Processing Systems}}},
  2016.

\bibitem[Titsias et~al.(2020)Titsias, Nikoloutsopoulos, and
  Galashov]{titsias2020information}
Michalis~K. Titsias, Sotirios Nikoloutsopoulos, and Alexandre Galashov.
\newblock Information {{Theoretic Meta Learning}} with {{Gaussian Processes}}.
\newblock \emph{arXiv:2009.03228 [cs, stat]}, 2020.

\bibitem[Tossou et~al.(2020)Tossou, Dura, Laviolette, Marchand, and
  Lacoste]{tossou2020adaptive}
Prudencio Tossou, Basile Dura, Francois Laviolette, Mario Marchand, and
  Alexandre Lacoste.
\newblock Adaptive {{Deep Kernel Learning}}.
\newblock \emph{arXiv:1905.12131 [cs, stat]}, 2020.

\bibitem[Varin et~al.(2011)Varin, Reid, and Firth]{varin2011overview}
Cristiano Varin, Nancy Reid, and David Firth.
\newblock An {{Overview Of Composite Likelihood Methods}}.
\newblock \emph{Institute of Statistical Science, Academia Sinica}, 2011.

\bibitem[Vinyals et~al.(2016)Vinyals, Blundell, Lillicrap, Kavukcuoglu, and
  Wierstra]{vinyals2016matching}
Oriol Vinyals, Charles Blundell, Timothy Lillicrap, Koray Kavukcuoglu, and Daan
  Wierstra.
\newblock Matching networks for one shot learning.
\newblock In D.~Lee, M.~Sugiyama, U.~Luxburg, I.~Guyon, and R.~Garnett,
  editors, \emph{Advances in Neural Information Processing Systems}, volume~29.
  {Curran Associates, Inc.}, 2016.

\bibitem[Wah et~al.(2011)Wah, Branson, Welinder, Perona, and
  Belongie]{wah2011caltechucsd}
Catherine Wah, Steve Branson, Peter Welinder, Pietro Perona, and Serge
  Belongie.
\newblock The {{Caltech}}-{{UCSD Birds}}-200-2011 {{Dataset}}.
\newblock Technical Report CNS-TR-2011-001, {California Institute of
  Technology}, 2011.

\bibitem[Wang et~al.(2019)Wang, Wang, and Truong]{wang2019customizable}
Kuan-Chieh Wang, Jixuan Wang, and Khai Truong.
\newblock Customizable {{Facial Gesture Recognition For Improved Assistive
  Technology}}.
\newblock In \emph{{{ICLR AI}} for {{Social Good Workshop}}}, 2019.

\bibitem[Wen et~al.(2018)Wen, Vicol, Ba, Tran, and Grosse]{wen2018flipout}
Yeming Wen, Paul Vicol, Jimmy Ba, Dustin Tran, and Roger Grosse.
\newblock Flipout: {{Efficient Pseudo}}-{{Independent Weight Perturbations}} on
  {{Mini}}-{{Batches}}.
\newblock In \emph{International {{Conference}} on {{Learning
  Representations}}}, 2018.

\bibitem[Williams and Barber(1998)]{williams1998bayesian}
Christopher K.~I. Williams and D.~Barber.
\newblock Bayesian classification with {{Gaussian}} processes.
\newblock \emph{IEEE Transactions on Pattern Analysis and Machine
  Intelligence}, 20\penalty0 (12):\penalty0 1342--1351, 1998.

\bibitem[Wilson et~al.(2016)Wilson, Hu, Salakhutdinov, and
  Xing]{wilson2016deep}
Andrew~Gordon Wilson, Zhiting Hu, Ruslan Salakhutdinov, and Eric~P. Xing.
\newblock Deep kernel learning.
\newblock In Arthur Gretton and Christian~C. Robert, editors, \emph{Proceedings
  of the 19th International Conference on Artificial Intelligence and
  Statistics}, volume~51 of \emph{Proceedings of Machine Learning Research},
  {Cadiz, Spain}, 2016. {PMLR}.

\bibitem[Windle et~al.(2014)Windle, Polson, and Scott]{windle2014sampling}
Jesse Windle, Nicholas~G. Polson, and James~G. Scott.
\newblock Sampling {{Polya}}-{{Gamma}} random variates: Alternate and
  approximate techniques.
\newblock \emph{arXiv:1405.0506 [stat]}, 2014.

\bibitem[Yoon et~al.(2018)Yoon, Kim, Dia, Kim, Bengio, and
  Ahn]{yoon2018bayesian}
Jaesik Yoon, Taesup Kim, Ousmane Dia, Sungwoong Kim, Yoshua Bengio, and Sungjin
  Ahn.
\newblock Bayesian {{Model}}-{{Agnostic Meta}}-{{Learning}}.
\newblock In \emph{Advances in {{Neural Information Processing Systems}}},
  2018.

\end{thebibliography}
\appendix
\section{Efficient Conditional Sampling of $\rvf$}\label{sec:efficient_sampling}

The Gibbs conditional distribution over $\rvf$ is given by:
\begin{equation}
  p(\rvf | \rmX, \rvy, \rvomega) = \mathcal{N}(\rvf | \tilde{\rmSigma}(\rmK^{-1} \rvmu + \rmA^\top \rvkappa), \tilde{\rmSigma}),  \label{eq:f_conditional_app}
\end{equation}
where $\tilde{\bm{\Sigma}} = (\rmK^{-1} + \rmA^\top \bm{\Omega} \rmA)^{-1}.$
Naively sampling from this distribution requires $\mathcal{O}(C^3 N^3)$
computation since $\tilde{\bm{\Sigma}}$ is a $CN \times CN$ matrix. Here we
describe a method for sampling from this distribution that requires
$\mathcal{O}(C N^3)$ computation instead.

First, we note that \eqref{eq:f_conditional_app} can be interpreted as
the conditional distribution $p(\rvf|\rvz = \rmOmega^{-1} \rvkappa)$ resulting
from the following marginal distribution $p(\rvf)$ and conditional
$p(\rvz|\rvf)$:
\begin{align}
  p(\rvf) &= \mathcal{N}(\rvf | \rvmu, \rmK) \\
  p(\rvz | \rvf) &= \mathcal{N}(\rvz | \rmA \rvf, \rmOmega^{-1}),
\end{align}
where we have made implicit the dependence on $\rmX, \rmY$, and $\rvomega$ for
brevity of notation. Equivalently, the distribution over $\rvf$ and $\rvz$ can
be represented by the partitioned Gaussian
\begin{align}
  \left[
  \begin{array}{c} \rvf \\ \rvz \end{array}
  \right] \sim \mathcal{N}\left(
  \left[ \begin{array}{c} \rvmu \\ \rmA \rvmu \end{array} \right],
  \left[
  \begin{array}{cc} \rmK &  \rmK \rmA^\top \\ \rmA \rmK & \rmA \rmK \rmA^\top + \rmOmega^{-1} \end{array}
                                                          \right] \right).
\end{align}
The conditional distribution $p(\rvf | \rvz)$ is given as:
\begin{equation}
  p(\rvf | \rvz) = \mathcal{N}(\rvf | \tilde{\rmSigma} ( \rmK^{-1} \rvmu + \rmA^\top \rmOmega \rvz), \tilde{\rmSigma}),
\end{equation}
where $\tilde{\rmSigma} = (\rmK^{-1} + \rmA^\top \rmOmega \rmA)^{-1}.$ Note that
$p(\rvf | \rvz = \rmOmega^{-1} \bm{\kappa})$ recovers our desired Gibbs
conditional distribution from \eqref{eq:f_conditional_app}.

An efficient approach to conditional Gaussian sampling is due to
\citet{hoffman1991constrained} and described in greater clarity
by~\citet{doucet2010note}. The procedure is as follows:

\begin{enumerate}
  \item Sample $\rvf_0 \sim p(\rvf)$ and $\rvz_0 \sim p(\rvz | \rvf)$.
  \item Return
        $\bar{\rvf} = \rvf_0 + \rmK \rmA^\top (\rmA \rmK \rmA^\top + \rmOmega^{-1})^{-1} (\rmOmega^{-1} \rvkappa - \rvz_0)$
        as the sample from $p(\rvf | \rvz)$.
\end{enumerate}
$\rmK$ is block diagonal and thus sampling from $p(\mathbf{f})$ requires
$\mathcal{O}(CN^3)$ time. $\rmA \rvf$ can be computed in $\mathcal{O}(CN)$ time,
since each entry is the difference between $f_i^{y_i}$ and $f_i^c$ for some $i$
and $c$. Overall, step 1 requires $\mathcal{O}(CN^3)$ time.

We now show how to compute $\bar{\rvf}$ from step 2 in $\mathcal{O}(CN^3)$ time.
We first expand $(\rmA \rmK \rmA^\top + \rmOmega^{-1})^{-1}$:
\begin{align}
  (\rmA \rmK \rmA^\top + \rmOmega^{-1})^{-1} = \rmOmega - \rmOmega \rmA (\rmK^{-1} + \rmA^\top \rmOmega \rmA)^{-1} \rmA^\top \rmOmega
\end{align}
We substitute into the expression for $\bar{\rvf}$:
\begin{align}
  \bar{\rvf} &= \rvf_0 + \rmK \rmA^\top (\rmOmega - \rmOmega \rmA (\rmK^{-1} + \rmA^\top \rmOmega \rmA)^{-1} \rmA^\top \rmOmega)  (\rmOmega^{-1} \rvkappa - \rvz_0) \\
             &= \rvf_0 + \rmK \rmA^\top \rmOmega (\rmOmega^{-1} \rvkappa - \rvz_0) - \rmK \rmA^\top \rmOmega \rmA (\rmK^{-1} + \rmA^\top \rmOmega \rmA)^{-1} \rmA^\top \rmOmega  (\rmOmega^{-1} \rvkappa - \rvz_0) \\
             &= \rvf_0 + \rmK  \rvv  - \rmK \rmA^\top \rmOmega \rmA (\rmK^{-1} + \rmA^\top \rmOmega \rmA)^{-1} \rvv, \label{eq:fbar_three_terms}
\end{align}
where we have defined
$\rvv \triangleq \rmA^\top \rmOmega (\rmOmega^{-1} \rvkappa - \rvz_0)$.

Now let
$\rvd \triangleq (d_1^1, \ldots, d_N^1, d_1^2, \ldots, d_N^2, \ldots, d_1^C, \ldots, d_N^C)^\top$,
where $d_i^c = Y_{ic} \sum_{c'} \omega_i^{c'}$. Define $\mathbf{Y}^\dag$ to be
the $CN \times N$ matrix produced by vertically stacking
$\text{diag}(Y_{\cdot c}),$ and let $\rmW^\dag$ be the $CN \times N$ matrix
produced by vertically stacking
$\text{diag}((\omega_1^c, \ldots, \omega_N^c)^\top)$. $\rmA^\top \rmOmega \rmA$
may then be written as follows:
\begin{align}
  \rmA^\top \rmOmega \rmA &= \rmD - \rmS \rmP \rmS^\top\text{, where} \label{eq:dsps} \\
  \rmD &= \rmOmega + \text{diag}(\rvd), \\
  \rmS &= \left[ \begin{array}{cc} \rmY^\dag & \rmW^\dag \end{array} \right], \\
  \rmP &= \left[ \begin{array}{cc} \mathbf{0}_N & \mathbf{I}_N \\
                   \mathbf{I}_N & \mathbf{0}_N
                 \end{array} \right].
\end{align}
Substituting \eqref{eq:dsps} into \eqref{eq:fbar_three_terms}:
\begin{align}
  \bar{\rvf} &= \rvf_0 + \rmK  \rvv  - \rmK \rmA^\top \rmOmega \rmA (\rmK^{-1} + \rmD - \rmS \rmP \rmS^\top)^{-1} \rvv. \label{eq:fbar_dsps}
\end{align}
Now we expand $(\rmK^{-1} + \rmD - \rmS \rmP \rmS^\top)^{-1}:$
\begin{align}
  (\rmK^{-1} + \rmD - \rmS \rmP \rmS^\top)^{-1} &=  \rmE - \rmE \rmS (\rmS^\top \rmE \rmS-\rmP^{-1})^{-1} \rmS^\top \rmE, \label{eq:expand}
\end{align}
where $\rmE = (\rmK^{-1} + \rmD)^{-1} = \rmK (\rmK + \rmD^{-1})^{-1} \rmD^{-1}$
is a block-diagonal matrix that can be computed in $\mathcal{O}(CN^3)$ time,
since $\rmD$ is diagonal and $\rmK$ is block diagonal. Now, substituting
\eqref{eq:expand} back into \eqref{eq:fbar_dsps},
\begin{align}
  \bar{\rvf} &= \rvf_0 + \rmK  \rvv  - \rmK \rmA^\top \rmOmega \rmA \rmE \rvv + \rmK \rmA^\top \rmOmega \rmA \rmE \rmS (\rmS^\top \rmE \rmS - \rmP^{-1})^{-1} \rmS^\top \rmE \rvv. \label{eq:efficient_expression}
\end{align}
Note that $(\rmS^\top \rmE \rmS - \rmP^{-1})^{-1}$ is a $2N \times 2N$ matrix
and thus can be inverted in $\mathcal{O}(N^3)$ time. The overall complexity is
therefore $\mathcal{O}(C N^3)$.

\section{Experimental Details}\label{sec:experimental_details}

Here we provide more details about our experimental setup for our classification
experiments, which are based on the protocol of \citep{patacchiola2020bayesian}.

\subsection{Datasets}
We used the four dataset scenarios described below. The first three are the same
used by \citet{chen2019closer} and the final was proposed by
\citet{patacchiola2020bayesian}.
\begin{itemize}
  \item \textbf{CUB.} Caltech-UCSD Birds (CUB) \citep{wah2011caltechucsd}
        consists of 200 classes and 11,788 images. A split of 100 training, 50
        validation, and 50 test classes was used \citep{hilliard2018fewshot,
        chen2019closer}.
  \item \textbf{mini-Imagenet.} The mini-Imagenet dataset
        \citep{vinyals2016matching} consists of 100 classes with 600 images per
        class. We used the split proposed by \citet{ravi2017optimization}, which
        has 64 classes for training, 16 for validation, and 20 for test.
  \item \textbf{mini-Imagenet$\rightarrow$CUB.} This cross-domain transfer
        scenario takes the training split of mini-Imagenet and the validation \&
        test splits of CUB.
  \item \textbf{Omniglot $\rightarrow$ EMNIST.} We use the same setup as
        proposed by \citet{patacchiola2020bayesian}. Omniglot
        \citep{lake2011one} consists of 1,623 classes, each with 20 examples,
        and is augmented by rotations of 90 degrees to create 6,492 classes, of
        which 4,114 are used for training. The EMNIST dataset
        \citep{cohen2017emnist}, consisting of 62 classes, is split into 31
        training and 31 test classes.
\end{itemize}

\subsection{Training Details} All methods employed the commonly-used Conv4
architecture \citep{vinyals2016matching} (see
Table~\ref{tab:conv4_architecture_all} for a detailed specification), except
ABML which used 32 filters throughout. All of our experiments used the Adam
\citep{kingma2015adam} optimizer with learning rate $10^{-3}$. During training,
all models used epochs consisting of 100 randomly sampled episodes. A single
gradient descent step on the encoder network and relevant hyperparameters is
made per episode. All 1-shot models are trained for 600 epochs and 5-shot models
are trained for 400 epochs. Each episode contained 5 classes (5-way) and 16
query examples. At test time, 15 query examples are used for each episode. Early
stopping was performed by monitoring accuracy on the validation set. The
validation set was not used for retraining.

We train both marginal likelihood and predictive likelihood versions of our
models. For P\'olya-Gamma sampling we use the PyP\'olyaGamma
package\footnote{\url{https://github.com/slinderman/pypolyagamma}}. During
training, we use a single step of Gibbs ($T$=1). For evaluation, we run until
$T=50$. In both training and evaluation, we use $M=20$ parallel Gibbs chains to
reduce variance.

\begin{table*}[!htb]
  \centering
  \caption{Specification of Conv4 architecture. \texttt{Conv2d} layers are
    $3 \times 3$ with stride $1$ and \texttt{same} padding. \texttt{MaxPool2d}
    layers are $2 \times 2$ with stride $2$ and \texttt{valid} padding.}
  \label{tab:conv4_architecture_all}
  \subtable[Omniglot$\rightarrow$EMNIST dataset.]{
    \label{tab:conv4s_architecture}
    \begin{tabular}{cl}
      \hline
      \textbf{Output Size} & \textbf{Layers} \\
      \hline
      1 $\times$ 28 $\times$ 28 & Input image \\ \hline
      64 $\times$ 14 $\times$ 14 & \begin{tabular}{@{}l@{}} \texttt{Conv2d} \\ \texttt{BatchNorm2d} \\ \texttt{ReLU} \\ \texttt{MaxPool2d}  \end{tabular} \\  \hline
      64 $\times$ 7 $\times$ 7 & \begin{tabular}{@{}l@{}} \texttt{Conv2d} \\ \texttt{BatchNorm2d} \\ \texttt{ReLU} \\ \texttt{MaxPool2d}    \end{tabular} \\ \hline
      64 $\times$ 3 $\times$ 3 & \begin{tabular}{@{}l@{}} \texttt{Conv2d} \\ \texttt{BatchNorm2d} \\ \texttt{ReLU} \\ \texttt{MaxPool2d}    \end{tabular} \\ \hline
      64 $\times$ 1 $\times$ 1 & \begin{tabular}{@{}l@{}} \texttt{Conv2d} \\ \texttt{BatchNorm2d} \\ \texttt{ReLU} \\ \texttt{MaxPool2d} \end{tabular} \\ \hline
      64 & \texttt{Flatten} \\
      \hline
    \end{tabular}
  } \hspace{1cm}\subtable[All other datasets.]{
    \label{tab:conv4_architecture}
    \begin{tabular}{cl}
      \hline
      \textbf{Output Size} & \textbf{Layers} \\
      \hline
      3 $\times$ 84 $\times$ 84 & Input image \\ \hline
      64 $\times$ 42 $\times$ 42 & \begin{tabular}{@{}l@{}} \texttt{Conv2d} \\ \texttt{BatchNorm2d} \\ \texttt{ReLU} \\ \texttt{MaxPool2d}  \end{tabular} \\  \hline
      64 $\times$ 21 $\times$ 21 & \begin{tabular}{@{}l@{}} \texttt{Conv2d} \\ \texttt{BatchNorm2d} \\ \texttt{ReLU} \\ \texttt{MaxPool2d}  \end{tabular} \\ \hline
      64 $\times$ 10 $\times$ 10 & \begin{tabular}{@{}l@{}} \texttt{Conv2d} \\ \texttt{BatchNorm2d} \\ \texttt{ReLU} \\ \texttt{MaxPool2d}  \end{tabular} \\ \hline
      64 $\times$ 5 $\times$ 5 & \begin{tabular}{@{}l@{}} \texttt{Conv2d} \\ \texttt{BatchNorm2d} \\ \texttt{ReLU} \\ \texttt{MaxPool2d}  \end{tabular} \\ \hline
      1600 & \texttt{Flatten} \\
      \hline
    \end{tabular}
  }
\end{table*}

\section{Effect of Kernel Choice on Classification Accuracy}\label{sec:kernel_choice}

In this section, we examine the effect of kernel choice on classification
accuracy for our proposed One-vs-Each P\'olya-Gamma OVE GPs.

\paragraph{Cosine Kernel.} In the main paper, we showed results for the
following kernel, which we refer to as the ``cosine'' kernel due to its
resemblance to cosine similarity:
\begin{equation}
  k^\text{cos}(\xx, \xx'; \btheta, \alpha) = \exp(\alpha) \frac{g_{\btheta}(\xx)^\top g_{\btheta}(\xx')}{\|g_{\btheta}(\xx)\| \|g_{\btheta}(\xx')\|},
\end{equation}
where $g_{\btheta}(\cdot)$ is a deep neural network that outputs a
fixed-dimensional encoded representation of the input and $\alpha$ is the scalar
log output scale. Both $\btheta$ and $\alpha$ are considered hyperparameters and
learned simultaneously as shown in Algorithm~\ref{alg:pg_learning}. We found
that this kernel works well for a range of datasets and shot settings. We note
that the use of cosine similarity is reminiscent of the approach taken by
Baseline$++$ method of \citep{chen2019closer}, which computes the softmax over
cosine similarity to class weights.

Here we consider three additional kernels: linear, RBF, and normalized RBF.
\paragraph{Linear Kernel.} The linear kernel is defined as follows:
\begin{equation}
  k^\text{lin}(\xx, \xx'; \btheta, \alpha) = \frac{1}{D} \exp(\alpha) g_{\btheta}(\xx)^\top g_{\btheta}(\xx'),
  \label{eq:linear_kernel}
\end{equation}
where $D$ is the output dimensionality of $g_{\btheta}(\xx)$. We apply this
dimensionality scaling because the dot product between $g_{\btheta}(\xx)$ and
$g_{\btheta}(\xx')$ may be large depending on $D$.

\paragraph{RBF Kernel.} The RBF (also known as squared exponential) kernel can
be defined as follows:
\begin{equation}
  k^\text{rbf}(\xx, \xx'; \btheta, \alpha, \ell) = \exp(\alpha) \exp \left(-\frac{1}{2D \exp(\ell)^2} \| g_{\btheta}(\xx) - g_{\btheta}(\xx') \|^2 \right),
\end{equation}
where $\ell$ is the log lengthscale parameter (as with $\alpha$, we learn the
$\ell$ alongside $\btheta$).

\paragraph{Normalized RBF Kernel.} Finally, we consider a normalized RBF kernel
similar in spirit to the cosine kernel:
\begin{equation}
  k^\text{rbf-norm}(\xx, \xx'; \btheta, \alpha, \ell) = \exp(\alpha) \exp \left(-\frac{1}{2 \exp(\ell)^2} \left\|\frac{g_{\btheta}(\xx)}{\| g_{\btheta}(\xx) \|} - \frac{g_{\btheta}(\xx')}{\| g_{\btheta}(\xx') \|}\right\|^2 \right).
\end{equation}

The results of our P\'olya-Gamma OVE GPs with different kernels can be found in
Tables \ref{tab_supplementary_classification_accuracy} and
\ref{tab_supplementary_crossdomain_accuracy}. In general, we find that the
cosine kernel works best overall, with the exception of
Omniglot$\rightarrow$EMNIST, where RBF does best.

\begin{table*}[hbtp]
  \centering
  \caption{Classification accuracy for Pólya-Gamma OVE GPs (our method) using
    different kernels. Cosine is overall the best, followed closely by linear.
    RBF-based kernels perform worse, except for the Omniglot$\rightarrow$EMNIST
    dataset. Evaluation is performed on 5 randomly generated sets of 600 test
    episodes. Standard deviation of the mean accuracy is also shown. ML =
    Marginal Likelihood, PL = Predictive Likelihood.}
  \vspace{0.1in}
  \begin{tabular}{llcccc}
    \hline
    \textbf{} & \textbf{} & \multicolumn{2}{c}{\textbf{CUB}} & \multicolumn{2}{c}{\textbf{mini-ImageNet}} \\
    \small{\textbf{Kernel}} & \small{\textbf{Objective}} & \textbf{1-shot} & \textbf{5-shot} & \textbf{1-shot} & \textbf{5-shot}\\
    \hline
    \small{\textbf{Cosine}} & \small{\textbf{ML}} & \textbf{63.98 $\pm$ 0.43} & 77.44 $\pm$ 0.18 & 50.02 $\pm$ 0.35 & 64.58 $\pm$ 0.31 \\
    \small{\textbf{Linear}} & \small{\textbf{ML}} & 62.48 $\pm$ 0.26 & 77.94 $\pm$ 0.21 & \textbf{50.81 $\pm$ 0.30} & 66.66 $\pm$ 0.45\\
    \small{\textbf{RBF}} & \small{\textbf{ML}} & 58.49 $\pm$ 0.40 & 75.50 $\pm$ 0.18 & 50.33 $\pm$ 0.26 & 64.62 $\pm$ 0.37 \\
    \small{\textbf{RBF (normalized)}} & \small{\textbf{ML}} & 62.75 $\pm$ 0.32 & 78.71 $\pm$ 0.08 & 50.26 $\pm$ 0.31 & 64.84 $\pm$ 0.39\\ \hline
    \small{\textbf{Cosine}} & \small{\textbf{PL}} & 60.11 $\pm$ 0.26 & \textbf{79.07 $\pm$ 0.05} & 48.00 $\pm$ 0.24 & \textbf{67.14 $\pm$ 0.23} \\
    \small{\textbf{Linear}} & \small{\textbf{PL}} & 60.44 $\pm$ 0.39 & 78.54 $\pm$ 0.19 & 47.29 $\pm$ 0.31 & 66.66 $\pm$ 0.36 \\
    \small{\textbf{RBF}} & \small{\textbf{PL}} & 56.18 $\pm$ 0.69 & 77.96 $\pm$ 0.19 & 48.06 $\pm$ 0.28 & 66.66 $\pm$ 0.39\\
    \small{\textbf{RBF (normalized)}} & \small{\textbf{PL}} & 59.78 $\pm$ 0.34 & 78.42 $\pm$ 0.13 & 47.51 $\pm$ 0.20 & 66.42 $\pm$ 0.36\\
    \hline
  \end{tabular}
  \label{tab_supplementary_classification_accuracy}
\end{table*}

\begin{table*}[hbtp]
  \centering
  \caption{Cross-domain classification accuracy for Pólya-Gamma OVE GPs (our
    method) using different kernels. The experimental setup is the same as Table
    \ref{tab_supplementary_classification_accuracy}.}
  \vspace{0.1in}
  \begin{tabular}{llcccc}
    \hline
    \textbf{} & \textbf{} & \multicolumn{2}{c}{\textbf{Omniglot}$\rightarrow$\textbf{EMNIST}} & \multicolumn{2}{c}{\textbf{mini-ImageNet}$\rightarrow$\textbf{CUB}} \\
    \small{\textbf{Kernel}} & \small{\textbf{Objective}} & \textbf{1-shot} & \textbf{5-shot} & \textbf{1-shot} & \textbf{5-shot}\\
    \hline
    \small{\textbf{Cosine}} & \small{\textbf{ML}} & 68.43  $\pm$ 0.67 & 86.22 $\pm$ 0.20 & \textbf{39.66 $\pm$ 0.18} & 55.71 $\pm$ 0.31 \\
    \small{\textbf{Linear}} & \small{\textbf{ML}} & 72.42 $\pm$ 0.49 & 88.27 $\pm$ 0.20 & 39.61 $\pm$ 0.19 & 55.07 $\pm$ 0.29 \\
    \small{\textbf{RBF}} & \small{\textbf{ML}} & \textbf{78.05 $\pm$ 0.38} & 88.98 $\pm$ 0.16 & 36.99 $\pm$ 0.07 & 51.75 $\pm$ 0.27 \\
    \small{\textbf{RBF (normalized)}} & \small{\textbf{ML}} & 75.51 $\pm$ 0.47 & 88.86 $\pm$ 0.16 & 38.42 $\pm$ 0.16 & 54.20 $\pm$ 0.13 \\
    \hline
    \small{\textbf{Cosine}} & \small{\textbf{PL}} & 77.00 $\pm$ 0.50 & 87.52 $\pm$ 0.19 & 37.49 $\pm$ 0.11 & \textbf{57.23 $\pm$ 0.31} \\
    \small{\textbf{Linear}} & \small{\textbf{PL}} & 75.87  $\pm$ 0.43 & 88.77 $\pm$ 0.10 & 36.83 $\pm$ 0.27 & 56.46 $\pm$ 0.22 \\
    \small{\textbf{RBF}} & \small{\textbf{PL}} & 74.62 $\pm$ 0.35 & \textbf{89.87 $\pm$ 0.13} & 35.06 $\pm$ 0.25 & 55.12 $\pm$ 0.21 \\
    \small{\textbf{RBF (normalized)}} & \small{\textbf{PL}} & 76.01 $\pm$ 0.31 & 89.42 $\pm$ 0.16 & 37.50 $\pm$ 0.28 & 56.80 $\pm$ 0.39 \\
    \hline
  \end{tabular}
  \label{tab_supplementary_crossdomain_accuracy}
\end{table*}

\section{Additional Calibration Results} \label{sec:additional_calibration}

\begin{figure}[hbtp]
  \begin{center}
    \centerline{\includegraphics[width=\textwidth]{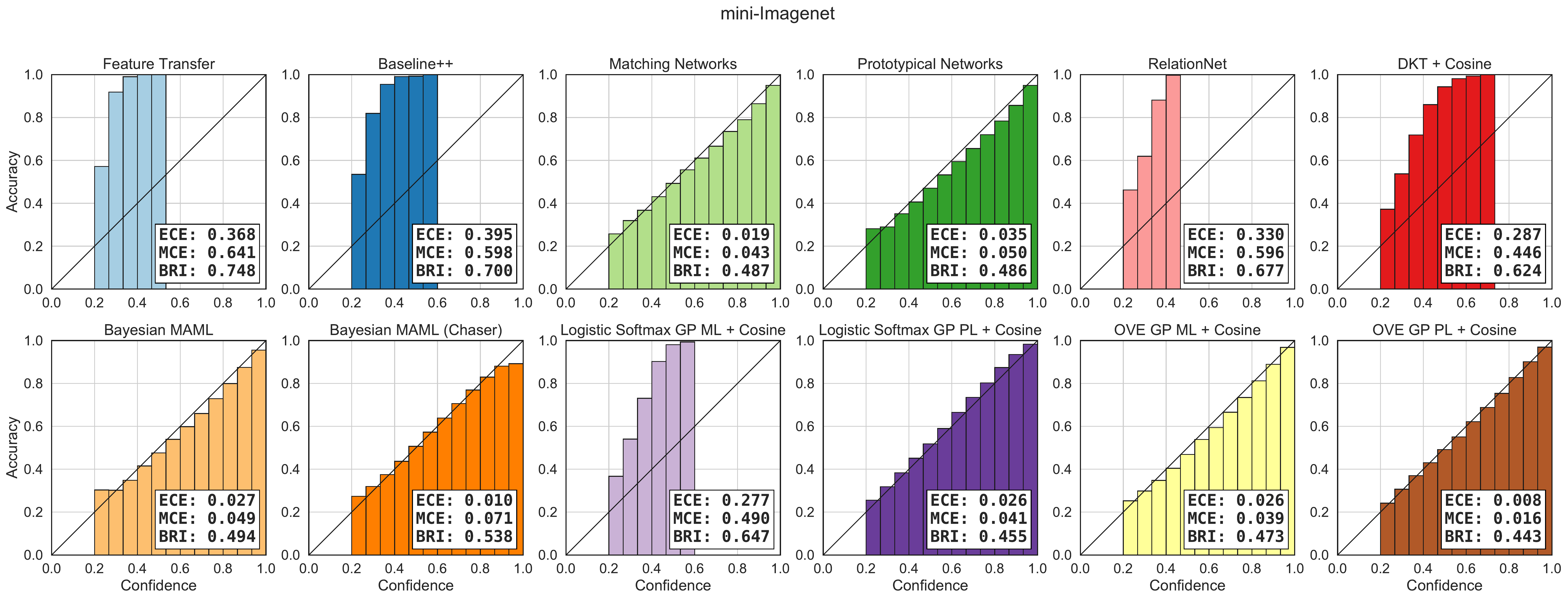}}
    \centerline{\includegraphics[width=\textwidth]{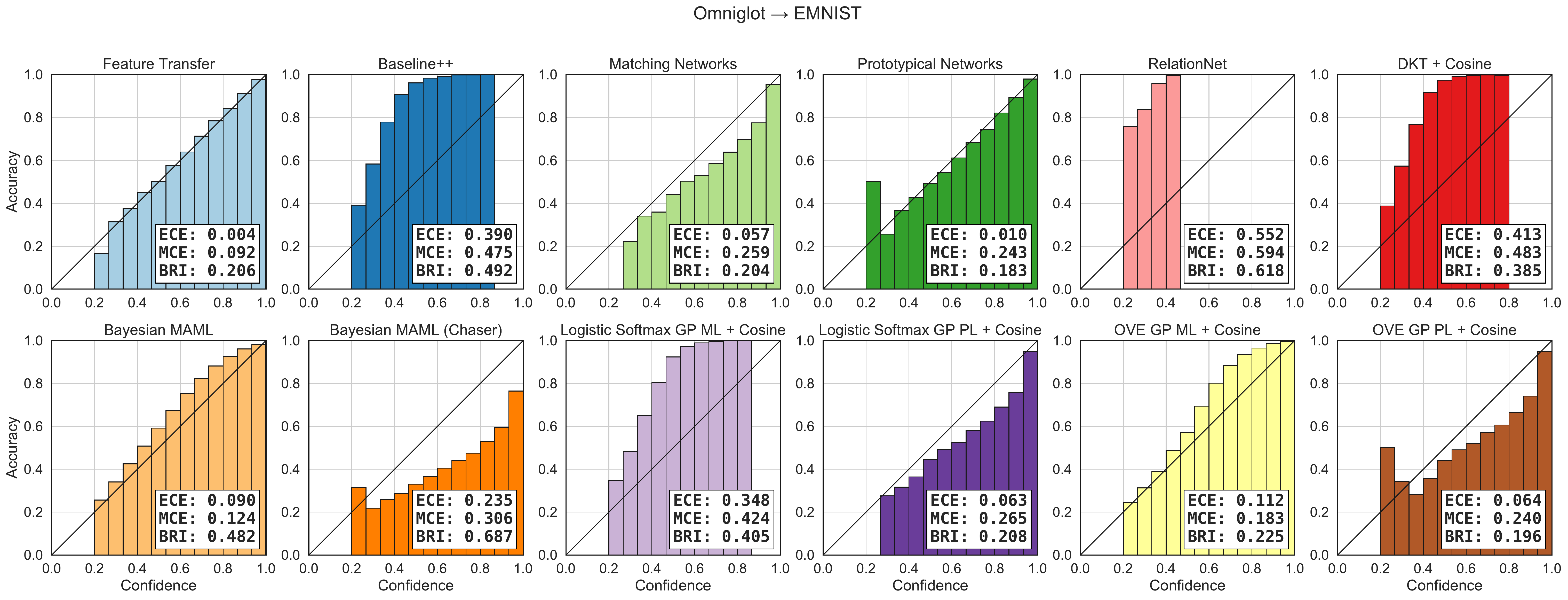}}
    \centerline{\includegraphics[width=\textwidth]{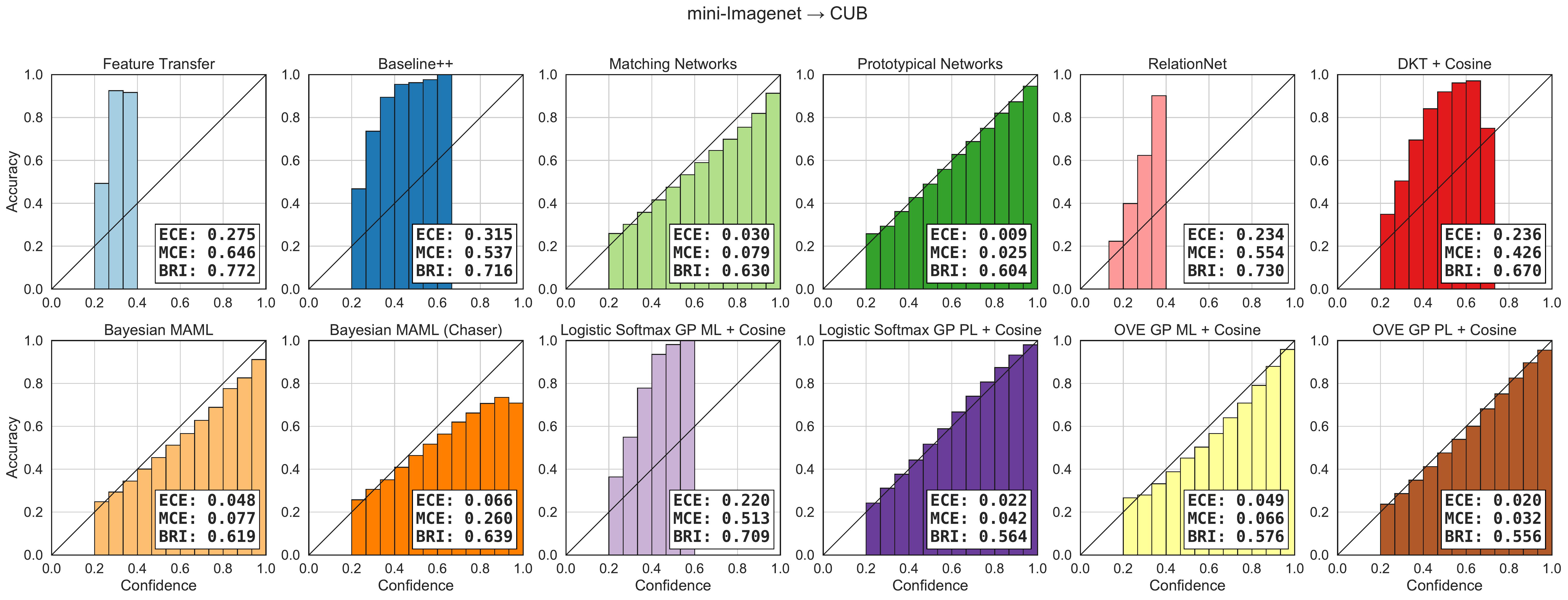}}
    \caption{Reliability diagrams, expected calibration error, maximum
      calibration error, and Brier scores for 5-shot 5-way tasks on
      mini-Imagenet, Omniglot$\rightarrow$EMNIST, and mini-Imagenet$\rightarrow$CUB. Metrics are computed on
      3,000 random tasks from the test
      set.} \label{fig:supplementary_calibration}
  \end{center}
  \vskip -0.4in
\end{figure}

In Figure \ref{fig:supplementary_calibration}, we include calibration results
for mini-Imagenet and Omniglot$\rightarrow$EMNIST. They follow similar trends to
the results presented in Section~\ref{sec:calibration_results}.

\clearpage
\section{Quantitative Robustness to Input Noise
  Results} \label{sec:additional_robustness}

In this section we include quantitative results for the robustness to input
noise results presented in Figure~\ref{fig:corruptions}. Results for Gaussian
noise are shown in Table~\ref{tab:gaussian_corruption_results}, impulse noise in
Table~\ref{tab:impulse_corruption_results}, and defocus blur in
Table~\ref{tab:defocus_corruption_results}.

\begin{table*}[!ht]
  \caption{Accuracy ($\%$) and Brier Score when applying Gaussian noise
    corruption of severity 5 to both the support and query set of test-time
    episodes. Results were evaluated across 1,000 randomly generated 5-shot
    5-way tasks.}
  \centering
  \begin{tabular}{l|cc|cc|cc}
    \hline
    \textbf{} & \multicolumn{2}{c}{\textbf{CUB}} & \multicolumn{2}{c}{\textbf{mini-ImageNet}} & \multicolumn{2}{c}{\textbf{mini-ImageNet}$\rightarrow$\textbf{CUB}} \\
    \small{\textbf{Method}} & \textbf{Acc. ($\uparrow$)} & \textbf{Brier ($\downarrow$)} & \textbf{Acc. ($\uparrow$)} & \textbf{Brier ($\downarrow$)} & \textbf{Acc. ($\uparrow$)} & \textbf{Brier ($\downarrow$)} \\
    \hline

    \small{\textbf{Feature Transfer}} &   30.45 &     0.775 &            22.58 &              0.799 &     22.75 &       0.799 \\
    \small{\textbf{Baseline$++$}} &   22.60 &     0.798 &            23.82 &              0.797 &     24.13 &       0.797 \\
    \small{\textbf{MatchingNet}} &   26.72 &     0.803 &            24.80 &              0.797 &     23.59 &       0.804 \\
    \small{\textbf{ProtoNet}} &   32.28 &     0.778 &            29.97 &              0.781 &     32.30 &       0.779 \\
    \small{\textbf{RelationNet}} &   25.23 &     0.799 &            23.69 &              0.800 &     20.00 &       0.800 \\
    \small{\textbf{DKT + Cosine}} &   29.54 &     0.779 &            27.78 &              0.792 &     31.94 &       0.782 \\
    \small{\textbf{Bayesian MAML}} &   22.79 &     0.905 &            20.52 &              0.963 &     20.46 &       0.949 \\
    \small{\textbf{Bayesian MAML (Chaser)}} &   20.20 &     1.133 &            20.41 &              1.118 &     21.39 &       1.039 \\
    \small{\textbf{LSM GP + Cosine (ML)}} &   27.92 &     0.787 &            22.43 &              0.798 &     22.36 &       0.799 \\
    \small{\textbf{LSM GP + Cosine (PL)}} &   31.21 &     0.772 &            31.77 &              0.768 &     \textbf{34.74} &       \textbf{0.754} \\
    \hline
    \small{\textbf{OVE PG GP + Cosine (ML)}} [ours] &   32.27 &     0.774 &            29.99 &              0.776 &     29.97 &       0.784 \\
    \small{\textbf{OVE PG GP + Cosine (PL)}} [ours] &   \textbf{33.01} &     \textbf{0.771} &            \textbf{33.29} &              \textbf{0.760} &     31.41 &       0.764 \\
    \hline
  \end{tabular}
  \label{tab:gaussian_corruption_results}
\end{table*}

\begin{table*}[!ht]
  \caption{Accuracy ($\%$) and Brier Score when applying impulse noise
    corruption of severity 5 to both the support and query set of test-time
    episodes. Results were evaluated across 1,000 randomly generated 5-shot
    5-way tasks.}
  \centering
  \begin{tabular}{l|cc|cc|cc}
    \hline
    \textbf{} & \multicolumn{2}{c}{\textbf{CUB}} & \multicolumn{2}{c}{\textbf{mini-ImageNet}} & \multicolumn{2}{c}{\textbf{mini-ImageNet}$\rightarrow$\textbf{CUB}} \\
    \small{\textbf{Method}} & \textbf{Acc. ($\uparrow$)} & \textbf{Brier ($\downarrow$)} & \textbf{Acc. ($\uparrow$)} & \textbf{Brier ($\downarrow$)} & \textbf{Acc. ($\uparrow$)} & \textbf{Brier ($\downarrow$)} \\
    \hline

    \small{\textbf{Feature Transfer}} &   30.20 &     0.776 &            23.54 &              0.798 &     22.87 &       0.799 \\
    \small{\textbf{Baseline$++$}} &   28.05 &     0.790 &            23.72 &              0.798 &     25.58 &       0.795 \\
    \small{\textbf{MatchingNet}} &   28.25 &     0.790 &            23.80 &              0.803 &     23.21 &       0.811 \\
    \small{\textbf{ProtoNet}} &   32.12 &     0.774 &            28.81 &              0.783 &     32.70 &       0.775 \\
    \small{\textbf{RelationNet}} &   25.23 &     0.799 &            23.13 &              0.800 &     20.00 &       0.800 \\
    \small{\textbf{DKT + Cosine}} &   29.74 &     0.778 &            29.11 &              0.789 &     32.26 &       0.781 \\
    \small{\textbf{Bayesian MAML}} &   22.76 &     0.903 &            20.50 &              0.970 &     20.56 &       0.950 \\
    \small{\textbf{Bayesian MAML (Chaser)}} &   20.25 &     1.172 &            20.51 &              1.116 &     21.45 &       1.022 \\
    \small{\textbf{LSM GP + Cosine (ML)}} &   28.18 &     0.787 &            21.82 &              0.799 &     23.64 &       0.797 \\
    \small{\textbf{LSM GP + Cosine (PL)}} &   32.10 &     \textbf{0.769} &            30.22 &              0.776 &     \textbf{35.09} &       \textbf{0.751} \\
    \hline
    \small{\textbf{OVE PG GP + Cosine (ML)}} [ours] &   31.41 &     0.778 &            29.66 &              0.778 &     30.28 &       0.783 \\
    \small{\textbf{OVE PG GP + Cosine (PL)}} [ours] &   \textbf{33.36} &     0.772 &            \textbf{33.23} &              \textbf{0.761} &     32.06 &       0.762 \\

    \hline
  \end{tabular}
  \label{tab:impulse_corruption_results}
\end{table*}

\begin{table*}[!ht]
  \caption{Accuracy ($\%$) and Brier Score when applying defocus blur corruption
    of severity 5 to both the support and query set of test-time episodes.
    Results were evaluated across 1,000 randomly generated 5-shot 5-way tasks.}
  \centering
  \begin{tabular}{l|cc|cc|cc}
    \hline
    \textbf{} & \multicolumn{2}{c}{\textbf{CUB}} & \multicolumn{2}{c}{\textbf{mini-ImageNet}} & \multicolumn{2}{c}{\textbf{mini-ImageNet}$\rightarrow$\textbf{CUB}} \\
    \small{\textbf{Method}} & \textbf{Acc. ($\uparrow$)} & \textbf{Brier ($\downarrow$)} & \textbf{Acc. ($\uparrow$)} & \textbf{Brier ($\downarrow$)} & \textbf{Acc. ($\uparrow$)} & \textbf{Brier ($\downarrow$)} \\
    \hline

    \small{\textbf{Feature Transfer}} &   38.03 &     0.734 &            33.06 &              0.791 &     33.47 &       0.792 \\
    \small{\textbf{Baseline$++$}} &   42.55 &     0.710 &            35.89 &              0.761 &     39.88 &       0.740 \\
    \small{\textbf{MatchingNet}} &   44.43 &     0.682 &            34.43 &              0.754 &     35.95 &       0.741 \\
    \small{\textbf{ProtoNet}} &   46.78 &     \textbf{0.676} &            36.92 &              \textbf{0.737} &     41.45 &       0.714 \\
    \small{\textbf{RelationNet}} &   40.81 &     0.759 &            30.11 &              0.790 &     25.69 &       0.794 \\
    \small{\textbf{DKT + Cosine}} &     45.34 &     0.695 &            38.29 &              \textbf{0.737} &     \textbf{45.17} &       \textbf{0.703} \\
    \small{\textbf{Bayesian MAML}} &   42.65 &     0.697 &            30.63 &              0.808 &     37.32 &       0.736 \\
    \small{\textbf{Bayesian MAML (Chaser)}} &   40.66 &     0.881 &            29.93 &              1.121 &     31.33 &       1.125 \\
    \small{\textbf{LSM GP + Cosine (ML)}} &   45.37 &     0.706 &            34.10 &              0.769 &     39.66 &       0.753 \\
    \small{\textbf{LSM GP + Cosine (PL)}} &   48.55 &     0.690 &            \textbf{39.46} &              \textbf{0.737} &     43.15 &       0.714 \\
    \hline
    \small{\textbf{OVE PG GP + Cosine (ML)}} [ours] &   46.46 &     0.701 &            37.65 &              0.775 &     43.48 &       0.723 \\
    \small{\textbf{OVE PG GP + Cosine (PL)}} [ours] &   \textbf{49.44} &     0.695 &            38.95 &              0.780 &     43.66 &       0.720 \\

    \hline
  \end{tabular}
  \label{tab:defocus_corruption_results}
\end{table*}

\end{document}